\documentclass{article}

\newif{\ifnonymous}\nonymoustrue
\usepackage[main,final]{neurips_2026} 

\usepackage[utf8]{inputenc}
\usepackage[T1]{fontenc}
\usepackage{hyperref}
\usepackage{url}
\usepackage{booktabs}
\usepackage{amsfonts}
\usepackage{nicefrac}
\usepackage{microtype}
\usepackage{xcolor}
\usepackage{graphicx}
\usepackage{makecell}

\title{Learning to Construct Practical Agentic Systems}

\author{
Aditya Kumar$^1$ \\
Carnegie Mellon University\\
\texttt{adityaku@andrew.cmu.edu}
\And
Zhihan Lei$^1$\\
Carnegie Mellon University\\
\texttt{lexl@andrew.cmu.edu}
\And
Jerry Yan$^1$ \\
Carnegie Mellon University\\
\texttt{jerryy2@cs.cmu.edu}
\AND
Joshua W. Momo \\
Carnegie Mellon University\\
\texttt{jmomo@andrew.cmu.edu}
\And Lauhitya Reddy\\
Dept.~of Computer Science\\
Emory University\\
\texttt{lreddy3@emory.cmu}
\And
Rafael Enrique Cabrera Jimenez \\
Carnegie Mellon University\\
\texttt{rafaelcabrerajimenez7@gmail.com}
\And 
Cassandra A. Cohen\\
Carnegie Mellon University\\
\texttt{ccohen2@andrew.cmu.edu}
\And
Arthur Kajiyama \\
Carnegie Mellon University\\
\texttt{akajiyam@andrew.cmu.edu }
\And
William W. Cohen\\
Carnegie Mellon University\\
\texttt{wcohen@cmu.edu}
}

\begin{document}

\maketitle

\begin{abstract}
Automated design and optimization of agentic LLM-based systems leads to sophisticated systems that substantially improve result
quality over off-the-shelf agentic patterns. However, studies of
fielded agentic systems show that production systems focus much more on issues such as simplicity, controllability,
and predictability of inference costs.  In this paper we propose
principled approaches to designing and optimizing \emph{practical}
agentic systems.  We describe an agent framework that enables
designers to enforce modularity in agentic systems, by defining
``pseudo-tools'' that call LLMs recursively on a
restricted context. Using this framework we hand-engineer agents for a diverse set of tasks, and show that relative to dynamically-planned workflows, hand-constructed
fixed workflows are generally cheaper and more accurate.  We then propose novel learning 
methods for the agentic components required by this framework, namely
pseudo-tools and fixed workflows. These learning methods generally outperform hand-engineered agents. We also exploit the modularity
of the framework to apply multi-objective optimization methods
to jointly optimize cost and response
quality and blend the results of multiple learning systems.
\end{abstract}

\section{Introduction}

\footnotetext[1]{Equal contribution.} \refstepcounter{footnote}

Automated design and optimization of agentic systems is an important research area, with much work on optimizing agentic systems \cite{khattab2023dspy, yuksekgonul2025optimizing, agrawal2026gepa,hu2024adas, gptswarm2024, zhang2025aflow}, and even developing self-improving ones \cite{zhang2026darwingodelmachineopenended,fang2025comprehensivesurveyselfevolvingai}. Although this has led to improvements in result quality, studies  suggest that different concerns dominate in production environments, such as simplicity and predictability \cite{pan2026measuringagentsproduction}.  For example, it was observed that agentic systems in use usually use a hand-coded static workflow, rather than a ReAct-style  \cite{yao2023ReAct} agent loop, and many use multiple LLMs for different tasks in the same workflow.  

In this paper, we develop \textbf{principled approaches to designing and optimizing agentic systems}, focusing on designing \textbf{practical systems that are well-suited to production settings}.  We believe this goal requires a \textbf{framework that enables designers to enforce modularity} in agentic systems. Our proposal is to do this by constructing ``pseudo-tools''---which look like external LLM tools, but call another LLM with a restricted context when they are called.

\textbf{Contributions}.  We begin by hand-engineering agents in the same framework for 19 diverse tasks, in domains including finance, math, planning, and health.   We verify that hand-constructed fixed workflows are indeed often faster and more accurate than dynamically-planned workflows.  We then introduce methods for improving and/or developing the components of these hand-constructed systems, namely, the workflows and the pseudo-tools invoked by workflows. Finally, we exploit the modularity of the framework to apply multiobjective optimization methods~\cite{nsga2} to jointly optimize cost and response quality.

\textbf{The proposed framework.}  When LLMs generate text, every token depends on all previously-generated tokens.  This is also true when LLMs are “thinking", or planning tool-using actions in a ReAct \cite{yao2023ReAct} loop (reason, act, repeat).  The high degree of interdependence in agentic decision-making and reasoning contrasts with traditional software systems, which are made up of modular pieces of software.

However, agents do have the ability to restrict context when invoking tools, as tools receive only a part of the context as input.  We propose to exploit this feature to modularize reasoning: specifically, we introduce “pseudo-tools" (ptools) which solve a subtask by recursively calling an LLM with a restricted context. 
Technically, the core unit of our framework is the \emph{interface}---a typed function stub with a natural-language docstring specifying intended behavior. An interface can be bound to different alternative \emph{implementations}: possible implementations would include an LLM call with a task-specific prompt; code generated on-the-fly given the input; a ReAct sub-agent; or static Python code.  For convenience in developing agents, there is also a default LLM-based implementation for every interface, which uses an LLM to predict the function stub's output given an input.  The binding between an interface and an implementation is easily reconfigured, and the binding for one interface can be selected independently of other binding choices. Because each interface exposes typed inputs and outputs, the space of valid implementations is constrained enough for tractable search, yet rich enough to span the full range from low-cost deterministic code to flexible but expensive multi-turn agent loops.
\section{Related work}

\textbf{Agent architectures.} Because LLMs require some external harness to call tools, and because LLMs can often be productively used in sequence, there are numerous frameworks for implementing agentic systems: well-known frameworks include LangChain/LangGraph, DSPy~\cite{khattab2023dspy}, Pydantic AI~\cite{pydanticai}, SmolAgents~\cite{smolagents}, and others~\cite{choure2025agentic}. Our framework has some novel elements, discussed below, but is very lightweight.\footnote{It has less than 1000 lines of Python in its core, and makes use of several existing packages as components: we use Pydantic AI's ReAct-style dynamic planner, SmolAgents' Python sandbox, and use LiteLLM~\cite{litellm} for LLM calls.}
Notably it was no explicit representation of workflows over LLM calls---instead workflows are implemented as Python functions.  (This is sometimes called a "native agent architecture").

\textbf{Agent optimization and workflow generation.} Agentic architectures are often designed for particular types of optimizations---e.g., DSPy supports prompt optimization well. 
The optimizations we focus on are inspired by work on workflow optimization and generation, such as ADAS and the later AFLOW \cite{hu2024adas,aflow2024}.  For instance, ADAS defines workflows (similarly to us) as code, and refines workflows by incrementally using a ``meta agent'' that is guided by a database of historically useful workflows; AFLOW represents agents as graphs, where each node is an LLM action, and edges encode data-flow dependencies and data transformations. In AFLOW the agent space is searched with Monte Carlo Tree Search (MCTS), rather than conducting a linear search as in ADAS. GPTSwarm \cite{zhuge2024gptswarm} treats agent optimization as graph optimization, using operators that alter the graph topology (e.g., by reducing the distance information flows), combined with node-level prompt optimization.
Most of these systems focus on improving quality along a single dimension, output quality, and often develop workflows that are complex and costly.  We focus here on automatically constructing workflows that are simple and efficient (as well as effective).

\textbf{Workflow Memory And Caching.}
An alternative to building workflows is to save and reuse workflows.  Agent Workflow Memory \cite{wang2024awm} stores complete trajectories
and retrieves them as in-context guides on new instances; similar systems include Agent Skill
Induction \cite{wang2025asi} and ExpeL \cite{zhao2024expel}.  
 These methods reuse
\emph{whole trajectories or skills}; we instead extract recurring
\emph{sub-steps} and promote them into typed pseudo-tools that the
agent composes through ordinary tool calls.

\section{Methods}

\subsection{Framework}

\textbf{Interfaces and implementations.} The agentic framework we use\ifnonymous\footnote{\url{https://github.com/wwcohen/secretagent}} \fi extends Pydantic AI \footnote{\url{https://pydantic.dev/docs/ai/overview/}} with two new concepts: \emph{interfaces} and \emph{implementations}.  An \emph{interface} looks like a Python function stub, and consists of a function name; a set of typed named inputs; an output type; and a natural language description of the behavior of the function (a Python ``docstring'').
An interface can be \emph{bound} to a type-compatible \emph{implementation}, and a bound interface can be called just like a Python function. The framework includes a set of pre-defined \emph{implementation factories} which can be used to produce implementations of various types:
ordinary Python functions; 
a single call to an LLM, using a \emph{default prompt} automatically derived from the interface description;
a single LLM call using a user-defined prompt template;\footnote{The template should have slots corresponding to the interfaces arguments} 
a ReAct agent plus a list of tools; 
a single LLM call that generates Python code (optionally using a given list of tools) appropriate for a given input, and then executes that code in a sandbox. 
There are also a small number of special-purpose compound implementations for common agentic tasks, e.g., combining an implementation with an interface that extracts an answer from an unstructured LLM response, or retrying an LLM call until a validation step succeeds. 

\textbf{The root interface and configurations.} One designated \emph{root interface} is called for the top-level input from a user or benchmark case.  A \emph{task configuration} is defined by the root interface, together with the set of bindings of every interface (including the root).  Task configurations are easy to change, serialize, and restore.

\textbf{Expressiveness.} This framework flexibly supports a broad set of agent behaviors.  Interfaces can act as tools, when bound to a Python implementation; as recursive LLM calls (pseudo-tools); or as subagents.  Alternative bindings for the root interface can also implement many standard agent behaviors: e.g., the root interface can be bound to a Python function that calls a mix of tools and pseudo-tools, providing a static, engineered workflow; alternatively, a workflow can be dynamically planned by binding the root interface to ReAct with the same set of (pseudo-)tools.  Many variations and interpolations between these behaviors can also be easily implemented, as we will discuss below.

\subsection{Learning and optimization methods}

The interface/implementation framework allows a designer to 
explore a broad space of possible agentic designs, which will lead to different trade-offs with respect to different dimensions of performance (e.g., cost, interpretability, output quality, etc).  Our goal is
to support manual exploration of this space and, where possible, to automate the exploration of this space.
Among the design decisions are: when should pseudo-tools be introduced to modularize agent behavior? given a toolkit, what kind of implementation is appropriate for the root interface---e.g., should actions be dynamically planned, or can a fixed workflow be used? how can the modularity of the system be exploited to improve efficiency or testability---e.g., can some pseudo-tools be implemented by smaller and cheaper LLMs, or even replaced with zero-cost Python code?  In this section we consider these decisions.

\subsubsection{Learning a pseudo-tool toolkit} \label{sec:learn-pseudo-toolkit}

One important design task is constructing pseudo-tools to modularize agent behavior.

The \emph{ptool inducer} learns a typed library of pseudo-tools from
recorded agent traces, and is itself a one-shot LLM pipeline rather
than a gradient-trained model.  The intuition is that even an
unstructured rollout---a ReAct trajectory, or a free-form
chain-of-thought---implicitly factors a task into a sequence of
reasoning operations, and that recurring operations can be promoted
into named, recallable subroutines for a future agent.

Concretely, the inducer takes as input a set of recorded rollouts
from a base agent on training instances.  We support two trace modes:
\emph{ReAct}, which extracts the per-step \texttt{thought} field of
each tool-using turn, and \emph{CoT}, which chunks the agent's
free-form chain-of-thought into atomic reasoning steps.  An optional
filter restricts induction to rollouts that produced a correct final
answer.

A four-stage LLM pipeline then converts these thoughts into ptool
stubs.  (1)~Each thought is labeled independently with a short,
case-independent action type (e.g., ``extract temporal constraints'',
``verify suspect alibi'').  (2)~If the resulting label set is large, a
merge call collapses synonymous categories into a smaller canonical
set.  (3)~Categories whose frequency exceeds a minimum count are kept
as candidates, and the top-$K$ are retained.  (4)~For each retained
category, a final LLM call synthesizes an interface stub---a name, a
typed signature, and a natural-language docstring that describes the
reasoning pattern as a self-contained subroutine.

The output is a Python module of typed function stubs, each bound to
a single-LLM-call implementation that uses the synthesized docstring
as the prompt; the module is then registered as the action space of
a downstream ReAct sub-agent.  At runtime each induced ptool reduces
to a single LLM call against its docstring; the downstream agent
never invokes the inducer or sees the synthesis prompts, so the
runtime cost is the same as an equivalently sized hand-written
toolkit.  Three hyperparameters control the pipeline---the maximum
number of ptools $K$, the minimum category count for retention, and
the correct-only flag, but they are fixed for all our experiments; for details see 
\S\ref{sec:learn-ptools-results}. 

\subsection{Distilling pseudo-tools to tools}

After decomposing a problem, it may be that some interfaces are simple enough to implement with small specialist LMs, or even without calling LLMs at all.  In particular, given collected target input/output examples and the interface ``stub'', strong coding models can sometimes generate Python code to implement some interface.


The \emph{code distillation learner} replaces a single pseudo-tool's LLM call with a Python function. Given an interface and the (input, output) pairs harvested from recorded rollouts (optionally restricted to rollouts that produced a correct final answer), the learner prompts a strong code-generating LLM for $n$ candidate functions, executes each on all the training cases, picks the best in a round by training accuracy, and feeds any failures back as targeted feedback for the next of $R$ rounds. The generated function is allowed to \emph{abstain} by returning \texttt{None} or raising an error, in which case the workflow backs off to the original LLM-based ptool; however, an error when the function does not abstain (an error of commission) cannot be recovered from.  The distilled code is therefore retained only when a holdout shows a low rate of errors of commission. 
The result of code distillation is \emph{ptool-distill}: a collection of low cost, code-based implementations that replace LLM calls for a subset of the steps.

\subsection{Learning a workflow} \label{sec:learn-workflow}

The experiments below will confirm the observations of  others \cite{pan2026measuringagentsproduction}, and show that well-designed static workflows are often faster and more accurate than ReAct-style agent loops.  This is no surprise: the flexibility of an agent loop comes at the price of unpredictability in cost and performance.  We thus explored two approaches to learning workflows for a task.

\textbf{Learning workflows by code distillation.}  This approach applies the  code distillation learner to the root interface itself, producing a Python workflow that orchestrates the existing toolkit. For this case, the code generation prompt is extended with the signatures and docstrings for every tool or ptool, a few sampled tool-call traces from successful rollouts of the existing workflow, and optional ReAct traces to show "thinking" episodes.  The prompt also includes some hand-written reference workflows from other benchmarks, so it is a $k$-shot rather than $0$-shot prompt. We instruct the learner that returning \texttt{None} is strictly preferred to an error of commission, as code distillation for workflows also allows backoff to a simulate call on the same root interface.


\textbf{Learning a workflow by hill-climbing.}
The \emph{orchestration learner} treats both the workflow and the ptool toolkit as editable Python source, and improves them jointly across iterations. The supervisor may rewrite the workflow body, refine ptool docstrings (which serve as their prompts under the default LLM-call binding), or override per-interface configuration such as backend model assignments. The learner takes as input the root interface, a training set, an eval set, and a starting workflow that is either (i) the benchmark's hand-written workflow or (ii) a workflow composed at iteration zero from the available ptools by a single LLM call. We call the former a workflow-seeded run and the latter a tool-seeded run. In a workflow-seeded run, the supervisor inherits both a workflow and a ptool toolkit, and may edit either; in a tool-seeded run, only the toolkit is given, and the initialization step described in Appendix~\ref{app:orchestrator-composer} synthesizes an iteration-zero workflow before the iterative loop begins. Each iteration evaluates the current workflow, samples failed cases, and prompts a strong \emph{supervisor} LLM with the current workflow source, a per-interface profiling summary, the failure traces, and the iteration history. The supervisor returns either edited source (optionally with configuration overrides, such as model assignments) or a no-change signal. A proposed edit is accepted if it improves training accuracy, or if it matches training accuracy and improves evaluation accuracy; otherwise is rejected. 
The learner terminates after a fixed iteration budget or after five consecutive non-improvements. The two modes let the same machinery serve two purposes: \emph{repairing} an existing workflow and \emph{inducing} a workflow from a given toolkit. Full algorithmic details and artifacts are in Appendix~\ref{app:orchestrator}.
  

\subsection{Bi-objective optimization to fuse alternatives}\label{sec:bi-objective}

\textbf{Choice of learners and implementations as discrete optimization.} When ptools are optimized by code distillation, the result may be a toolkit that is faster on average, but possibly less accurate (due to errors of commission), so choosing the best set of ptool implementations can be non-obvious---especially since some workflows might be more robust to tool errors than others. Tradeoff arises because LLM-calling ptools might require strong LLMs, or might be nearly as effective with weaker, cheaper LLMs.  In our architecture, all of these design choices can be expressed configuration choices, specifically choices about how to bind interfaces, and how to parameterize individual ptools (e.g., with ptool-specific model choices).  Alternative procedures for learning an agent, or revising an initial seed workflow and/or set of tools, can also be expressed as configuration choices. This leads to the question of how to best search a space of alternative configurations---in general, for a desirable tradeoff of output quality with other metrics such as cost, latency, or interpretability.



We use the DEAP framework~\cite{JMLR:v13:fortin12a} to perform multi-objective search over the  configuration space.
Concretely, each interface in the system contributes one or more \emph{genes} to a chromosome: a \emph{method gene} which selects among a discrete set of
available implementations, and a \emph{model gene} which selects from a discrete set of backend LLMs. We also allow higher-level genes which \emph{compound}---e.g., a single gene value may expand to multiple configuration overrides (e.g., selecting ``program-of-thought'' also sets tool
lists and token limits). Interface-level genes allow the optimizer to assign different models and methods to different stages of a workflow independently.
The search space is defined declaratively in YAML for each benchmark, based on available options (potentially including learned implementations).
The optimizer evaluates configurations by running each on validation data, and recording output correctness and LLM cost per query.  \footnote{In the experiments below, we used the NSGA-II algorithm~\cite{nsga2}, a population-based evolutionary
strategy that maintains a Pareto frontier over these two objectives.
NSGA-II uses uniform crossover and random-reset mutation over the categorical chromosome encoding, with crowding-distance-based tournament selection
to preserve diversity along the frontier.} 

As well as comparing learned implementations to hand-engineered ones, this formalization naturally supports heterogeneous LLM assignment: different sub-interfaces can be bound to different backend models, e.g., allowing the
optimizer to discover configurations where a cheap model handles extraction while an expensive model handles reasoning. 

\textbf{Caching.} The modularity of interfaces means that similar chromosomes often invoke the same interfaces with the same inputs; we thus cache LLM calls during optimization, storing both the output and the reported cost so that optimization is cheaper and faster while still reflecting the true cost of each
configuration.\footnote{Specifically, we cache the LLM output as well as the cost of the call via LiteLLM's built-in cache, so that optimization will be cheaper and faster, but still reflect true costs.  See Appendix~\ref{app:opt-cache} for a measurement of cache effectiveness on one of our sweeps.}  Caching can reduce the real cost of optimization sweeps by up to a factor of 8.

\section{Experimental results}

\subsection{Why are static workflows preferred in practice?}

To our knowledge, there are no published data on the outcome of systematic engineering of agentic systems, across multiple benchmarks, using a common experimental harness. To address this shortage, over the course of approximately six weeks, each coauthor implemented and optimized\footnote{Optimization was by "graduate student descent" on validation data, exploring different toolkits/ptoolkits, agent loops, and prompts. Most coauthors are future or current master's students in AI.} 1-3 benchmarks, while the group also engaged in literature review and refining the common framework.  Benchmarks selected were cited by related papers, or based on coauthor expertise and interest.

   \begin{table*}[ht]
\centering
\footnotesize
\caption{Left, result quality (generally accuracy) across multiple benchmarks for different root interface bindings.  
Right, cost in US dollars per 100 task instances using Together.ai's costs for DeepSeek-V3.1.
The static workflows are hand-engineered on a dev set.  The dynamic workflow (ReAct) uses the same pseudo-tools and tools. The zero-shot baseline uses the same prompt as ReAct but no tools, so makes only one LLM call.  DeepSeek-V3.1 was used as the LLM model throughout.} 
\label{tab:hero}
\begin{tabular}{lccc|ccc}
\toprule
 & \multicolumn{3}{c}{Correctness} & \multicolumn{3}{c}{Cost (per 100 examples)} \\
\cmidrule(lr){2-4} \cmidrule(lr){5-7}
Task & \makecell{Static\\Workflow} & \makecell{Dynamic\\Workflow\\(ReAct)} & \makecell{Zero-shot\\(Default\\Imp.)} & \makecell{Static\\Workflow} & \makecell{Dynamic\\Workflow\\(ReAct)} & \makecell{Zero-shot\\(Default\\Imp.)} \\
\midrule
BBH Date Understanding & $0.84$ & $\mathbf{0.93}$ & $0.52$ & $0.28$ & $0.99$ & $\mathbf{0.10}$ \\
BBH Geometric Shapes & $0.30$ & $0.35$ & $\mathbf{0.53}$ & $0.63$ & $2.40$ & $\mathbf{0.07}$ \\
BBH Penguins in a Table & $0.63$ & $0.72$ & $\mathbf{0.93}$ & $0.21$ & $0.54$ & $\mathbf{0.05}$ \\
BBH Sports Understanding & $0.87$ & $\mathbf{0.88}$ & $0.65$ & $0.13$ & $0.29$ & $\mathbf{0.02}$ \\
FinQA & $\mathbf{0.75}$ & $0.32$ & $0.62$ & $\mathbf{0.12}$ & $0.95$ & $0.14$ \\
MedAgentBench & $\mathbf{0.87}$ & $0.69$ & $0.00$ & $0.64$ & $0.65$ & $\mathbf{0.22}$ \\
MedCalc Formulas & $0.81$ & $\mathbf{0.82}$ & $0.60$ & $\mathbf{0.15}$ & $1.20$ & $0.19$ \\
MedCalc Rules & $\mathbf{0.50}$ & $0.47$ & $0.45$ & $0.27$ & $1.29$ & $\mathbf{0.21}$ \\
MUSR Murder & $\mathbf{0.68}$ & $0.62$ & $0.52$ & $0.47$ & $1.34$ & $\mathbf{0.11}$ \\
MUSR Objects & $\mathbf{0.58}$ & $0.36$ & $0.35$ & $0.34$ & $0.99$ & $\mathbf{0.10}$ \\
MUSR Teams & $\mathbf{0.61}$ & $0.49$ & $0.59$ & $0.28$ & $1.38$ & $\mathbf{0.14}$ \\
NaturalPlan Calendar & $\mathbf{0.62}$ & $0.46$ & $0.54$ & $0.46$ & $0.79$ & $\mathbf{0.09}$ \\
NaturalPlan Meeting & $\mathbf{0.37}$ & $0.23$ & $0.23$ & $0.57$ & $0.91$ & $\mathbf{0.11}$ \\
NaturalPlan Trip & $\mathbf{0.21}$ & $0.16$ & $0.17$ & $0.37$ & $1.19$ & $\mathbf{0.08}$ \\
Rulearena Airlines & $\mathbf{0.98}$ & $0.89$ & $0.41$ & $\mathbf{0.97}$ & $2.40$ & $1.12$ \\
Rulearena Tax & $\mathbf{0.50}$ & $0.15$ & $0.11$ & $1.01$ & $3.49$ & $\mathbf{0.68}$ \\
Rulearena NBA & $0.61$ & $0.59$ & $\mathbf{0.67}$ & $1.49$ & $2.95$ & $\mathbf{1.42}$ \\
Tabular Math WP & $0.90$ & $\mathbf{0.97}$ & $0.87$ & $0.07$ & $0.69$ & $\mathbf{0.04}$ \\
$\tau$ Bench Retail & $0.52$ & $\mathbf{0.56}$ & $0.11$ & $4.75$ & $7.03$ & $\mathbf{3.50}$ \\
\midrule
\textrm{Average} & $\mathbf{0.64}$ & $0.56$ & $0.47$ & $0.70$ & $1.66$ & $\mathbf{0.44}$ \\
\bottomrule
\end{tabular}

\end{table*}

\begin{figure}[b]
\centering
\includegraphics[width=0.4\textwidth]{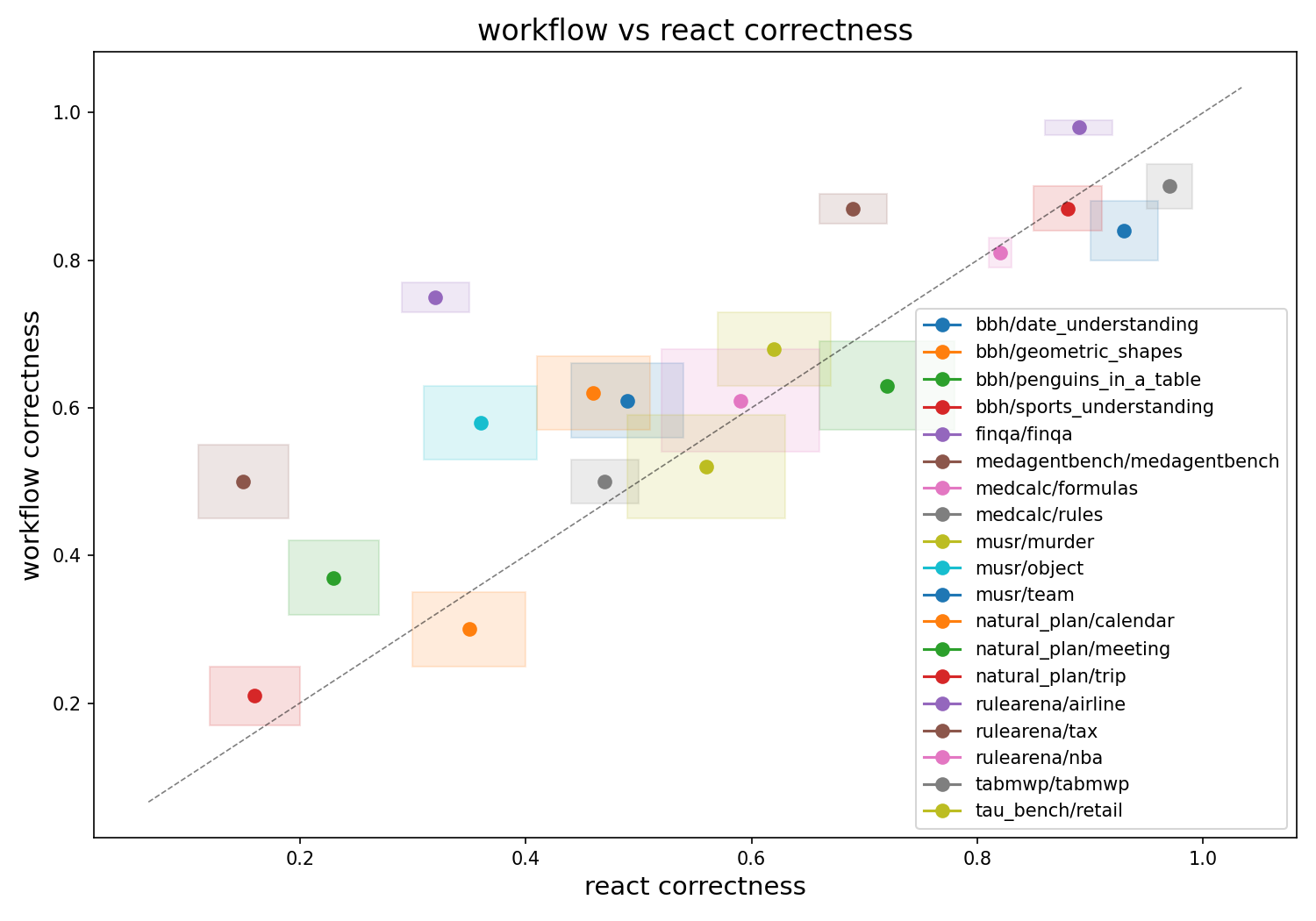}
\includegraphics[width=0.4\textwidth]{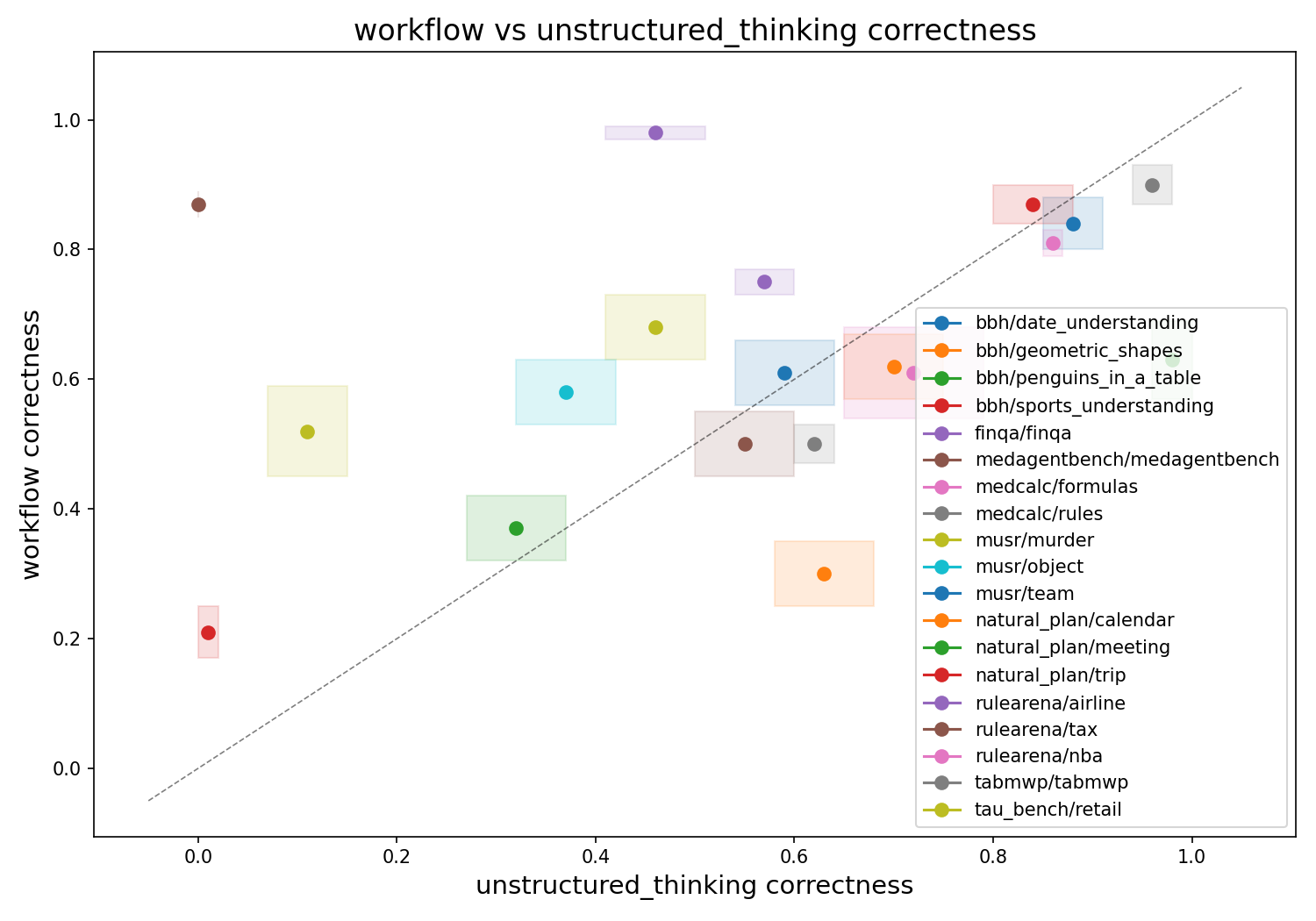}
\caption{\small Left, the accuracy of engineered static workflows compared to ReAct; right, accuracy of workflows compared to a traditional zero-shot tuned prompt.  Each box is a benchmark, where the y-axis position is mean workflow accuracy and the x-axis position is ReAct, so boxes above the line $y=x$ show that workflow performs best.  Box width is the standard error of the mean.} \label{fig:workflow}
\end{figure}

\textbf{Static workflows and pseudo-tools are fast and effective.}
In Table~\ref{tab:hero}) we show the results of static workflows across all benchmarks (on test data heldout during this optimization phase). We also show two ablations.  One uses a ReAct loop over the same toolkit (with a default prompt constructed from the interface description). This conditions ablates exactly the information present in the hand-engineered workflow, but retains the engineered set of tools/ptools.  The other is a zero-shot baseline, using the same top-level prompt as the ReAct loop, which ablates the engineered toolkit as well.

In Table~\ref{tab:hero} the engineered static workflows usually perform best. The left of Figure~\ref{fig:workflow} gives a graphic view of the comparison to ReAct, and also indicates the standard error of these measurements.  In Table~\ref{tab:hero} we also record the cost of each strategy, showing that static workflows are also cheaper than ReAct by a factor of more than 2 on average.

\textbf{The "agentic tax" is real.}  Dynamic workflows have an obvious computational cost compared to a fixed workflow. In these experiments they were also often fragile: e.g., tool/ptool calls often fail because of syntax or semantic errors. Hand-tuning can reduce these errors, but tuning dynamic agents can be expensive and slow.\footnote{Some poorly planned experiments with ReAct on RuleArena tasks with variant ptools cost hundreds of dollars.} 

\textbf{Single-shot LLMs are strong when Python tools are not needed.}
On the right in Figure~\ref{fig:workflow} (and in more detail in Appendix~\ref{app:hero}) we also compare to easy-to-tune manually constructed traditional zero-shot prompts. Briefly, the average correctness of these prompts is 0.56, the same as ReAct,\footnote{The 0.56 is noticibly higher than the default zero-shot prompt used in our ablations, which averaged 0.47.  Much of the improvement is from instructions which suggest workflow like-procedures to follow, which led to large gains on some tasks; if these are disallowed then the average score of traditional zero-shot prompts drops to 0.50.} but there are several cases where they outperform  agentic models.  However, there are benchmarks where "real" tools are absolutely necessary for performance (e.g., MedAgentBench and $\tau$ Bench).

\textbf{Engineered agentic systems follow many design patterns.}  The ability to easily design and invoke single-shot LLM tools is one of the goals and advantages of the ptool architecture, and it was used in many ways.  For example, in $\tau$-bench, a LLM-using ptool was be used to route between variant workflows, but calling it and conditioning on the output.  A frequently-used pattern to gain efficiency (not discussed to our knowledge in the literature) was to replace selected pseudo-tools for Pythonic "real" tools, which led to very low costs on some of the tasks.

\begin{table*}[tb]
\centering
\caption{Correctness for selected benchmarks using learned components. Lines 1--2 are the static workflow and ReAct results from Table~\ref{tab:hero}, shown as baselines. Orch-WfSeed starts from the hand-engineered workflow; Orch-ToolSeed starts from a composer-synthesized workflow over the toolkit (engineered or induced). All Orchestrate numbers are eval-split accuracy at the best-eval iteration over a 5-iteration run; see Appendix~D.5 for full sweep.}
\label{tab:learning}
\setlength{\tabcolsep}{3.5pt}
\renewcommand{\arraystretch}{1.05}
\footnotesize
\begin{tabular}{@{}lccccccccc@{}}
\toprule
 & \multicolumn{3}{c}{MuSR} & \multicolumn{2}{c}{NaturalPlan} & \multicolumn{1}{c}{RuleArena} & \multicolumn{2}{c}{MedCalc} & \multicolumn{1}{c}{} \\
\cmidrule(lr){2-4}
\cmidrule(lr){5-6}
\cmidrule(lr){7-7}
\cmidrule(lr){8-9}
Workflow/Ptools & Murder & Object & Team & Meeting & Trip & NBA & Formulas & Rules & Avg. \\
\midrule
human/human            & $0.68$ & $0.58$ & $0.61$ & $0.37$ & $0.21$ & $0.61$ & $0.81$ & $0.50$ & $0.55$ \\
ReAct/human            & $0.48$ & $0.34$ & $0.47$ & $0.29$ & $0.16$ & $0.50$ & $0.82$ & $0.47$ & $0.44$ \\ \midrule
ReAct/learned          & $0.75$ & $0.69$ & $0.68$ & $0.35$ & $0.01$ & $0.72$ & $0.59$ & $0.50$ & $0.54$ \\ \midrule
CodeDist/human         & $0.64$ & $0.61$ & $0.67$ & $0.27$ & $0.97$ & $\mathbf{1.00}$ & $0.70$ & $0.50$ & $0.67$ \\
CodeDist/learned       & $0.69$ & $0.48$ & $0.67$ & $\mathbf{1.00}$ & $0.78$ & $0.72$ & $0.54$ & $0.35$ & $0.65$ \\ \midrule
Orch-WfSeed/human      & $0.68$ & $0.64$ & $0.70$ & $\mathbf{1.00}$ & $0.11$ & $0.52$ & $0.78$ & $0.51$ & $0.62$ \\
Orch-ToolSeed/human    & $\mathbf{0.76}$ & $0.69$ & $0.74$ & $0.75$ & $\mathbf{1.00}$ & $0.26$ & $\mathbf{0.80}$ & $\mathbf{0.52}$ & $0.69$ \\
Orch-ToolSeed/learned  & $0.54$ & $\mathbf{0.71}$ & $\mathbf{0.80}$ & $\mathbf{1.00}$ & $\mathbf{1.00}$ & $0.74$ & $\mathbf{0.80}$ & $0.51$ & $\mathbf{0.76}$ \\
\bottomrule
\end{tabular}
\end{table*}

\subsection{Can pseudo-tools be learned?}\label{sec:learn-ptools-results}

The engineering done for the results of Table~\ref{tab:hero} requires designing two artifacts: a \emph{toolkit}, i.e., a set of tools and pseudo-tools, and a \emph{workflow}.  We consider here the problem of learning
a toolkit.

Many tasks clearly require external tools (e.g., for database access, calculation, etc), and use of these tools helps decompose the task into steps.  Here we focus on designing \emph{pseudo-tools} that decompose a \emph{reasoning process} into reusable, modular pieces.  We conjecture that techniques developed for this difficult induction process will be adaptable to tasks where external tools also exist, and selected several subtasks that seemed appropriate test cases.\footnote{We omitted subtasks which seemed to be solvable well without decomposition, some costly subtasks, and subtasks requiring external tools.}

For each of these tasks, we used the method of \S\ref{sec:learn-pseudo-toolkit} to induce a set of $K=5$ ptools, and used that toolkit with a ReAct loop.  The results are shown in row 3 of Table~\ref{tab:learning},
along with two baselines: row 1, the original human-engineered static workflow and human-engineered toolkit; and row 2, the same ReAct loop
with the human-engineered toolkit.  For these experiments, DeepSeek-V3~\cite{deepseekv3} was used both for learning the ptools and as the inference LLM for ReAct.

\textbf{Induced Pseudo-tools Perform Well.}  On average, the learned ptools outperform the engineered workflow and toolkit.  The improvement is consistent, with induced tools scoring better on 4/6 tasks.  More importantly, when the same workflow process (ReAct) is used, the induced ptools perform better on 5/6 tasks. We also evaluated induced ptools  in conjunction with the workflow learning methods described in \S\ref{sec:learn-workflow} (see below). Overall, the induced ptools also perform better overall than engineered ones in these conditions.  For more detail see Appendix~\ref{app:ptool-ind-exp}.

\begin{table*}[ht]
\centering
\caption{Cost (USD per 100 examples) for selected benchmarks using learned components. The first two lines are duplicated from the static workflow and ReAct results of Table~\ref{tab:hero}, and are shown here as baselines. Zeros indicate workflows that are pure Python with no LLM calls, which are considered zero-cost in our experiments.}
\label{tab:learning-cost}
\setlength{\tabcolsep}{3.5pt}
\renewcommand{\arraystretch}{1.05}
\footnotesize
\begin{tabular}{@{}lccccccccc@{}}
\toprule
 & \multicolumn{3}{c}{MuSR} & \multicolumn{2}{c}{NaturalPlan} & \multicolumn{1}{c}{RuleArena} & \multicolumn{2}{c}{MedCalc} & \multicolumn{1}{c}{} \\
\cmidrule(lr){2-4}
\cmidrule(lr){5-6}
\cmidrule(lr){7-7}
\cmidrule(lr){8-9}
Workflow/Ptools & Murder & Object & Team & Meeting & Trip & NBA & Formulas & Rules & Avg. \\
\midrule
human/human            & $0.47$ & $0.34$ & $\mathbf{0.28}$ & $0.57$ & $0.37$ & $1.49$ & $0.15$ & $\mathbf{0.27}$ & $0.49$ \\
ReAct/human            & $0.71$ & $0.53$ & $0.36$ & $0.94$ & $1.07$ & $3.09$ & $1.20$ & $1.29$ & $1.15$ \\
ReAct/learned          & $5.24$ & $2.36$ & $2.84$ & $6.56$ & $2.35$ & $68.40$ & $0.73$ & $0.87$ & $11.17$ \\
CodeDist/human         & $0.73$ & $0.88$ & $0.47$ & $0.68$ & $0.00$ & $0.00$ & $0.25$ & $0.38$ & $0.57$ \\
CodeDist/learned       & $0.83$ & $\mathbf{0.20}$ & $0.42$ & $0.00$ & $\mathbf{0.03}$ & $2.88$ & $0.20$ & $\mathbf{0.13}$ & $0.67$ \\
Orch-WfSeed/human      & $0.46$ & $0.78$ & $0.32$ & $0.00$ & $0.39$ & $1.50$ & $0.15$ & $0.28$ & $0.55$ \\
Orch-ToolSeed/human    & $0.49$ & $0.24$ & $0.39$ & $0.97$ & $0.00$ & $\mathbf{0.05}$ & $\mathbf{0.14}$ & $0.29$ & $\mathbf{0.37}$ \\
Orch-ToolSeed/learned  & $\mathbf{0.21}$ & $0.70$ & $1.03$ & $0.00$ & $0.00$ & $1.44$ & $0.55$ & $0.47$ & $0.73$ \\
\bottomrule
\end{tabular}
\end{table*}

\subsection{Can static workflows be learned?}\label{sec:learn-workflow-results}

We applied the two workflow-learning methods described in \S\ref{sec:learn-workflow}.  The code distillation learner was applied to the root interface, producing a Python workflow over the existing toolkit. The orchestration learner was run in both of its modes: workflow-seeded, starting from the hand-engineered workflow over the engineered toolkit; and tool-seeded, in which only a toolkit is provided. Workflow learning requires a strong coding model, so all orchestration-learner runs used Gemini 3.1 Pro~\cite{gemini31pro} as supervisor.

\textbf{Learned workflows outperform engineered ones.} The results are shown in the final four lines of Table~\ref{tab:learning}.  Lines 4 and 6 are using engineered toolkits with learned static workflows.  Learned workflows fairly consistently outperform the engineered ones (there are two exceptions for the code distillation learner, and one for the orchestration learner).  As noted above, using induced ptools gives another improvement on average.

\textbf{The "agentic tax" is real.} Table~\ref{tab:learning-cost} summarizes the cost of the various learned models, and emphasizes one of the practical advantages of using static workflows: their relative efficiency over dynamic, ReAct-style workflows.  The ptool induction method can produce tools that are effective but expensive to use in inference, but switching to a static workflow can dramatically reduce this overhead.\footnote{In the RuleArena/NBA task, the reduction in cost is nearly a factor of 50, and in MuSR/Murder it is a factor of 25.}

\textbf{Learning by coding is powerful but susceptible to "reward hacking".} Three of the tasks we selected turned out to be possible to implement without LLM calls at all--and both of our workflow learners discovered this, generating Python "workflows" sometimes hundreds of steps long that made no use of the toolkit at all, and instead combined regex-based extraction from inputs with search methods to find solutions.  Surprisingly, most of these generated solutions generalized well.\footnote{Many LLM benchmark problems are problems that are hard for LLM reasoners, but not intrinsically hard to solve computationally, and unless the inputs are linguistically complex coded shortcuts are possible.}  

We discuss this issue more in \S\ref{sec:limitations} but note here that the experimental results for our learning methods are qualitatively the same on the MuSR problems, where no "reward hacking" was observed.



\subsection{Can modularity be exploited to improve performance in other ways?}\label{sec:opt-results}


\begin{figure*}[tb]
\centering
\includegraphics[width=0.8\textwidth]{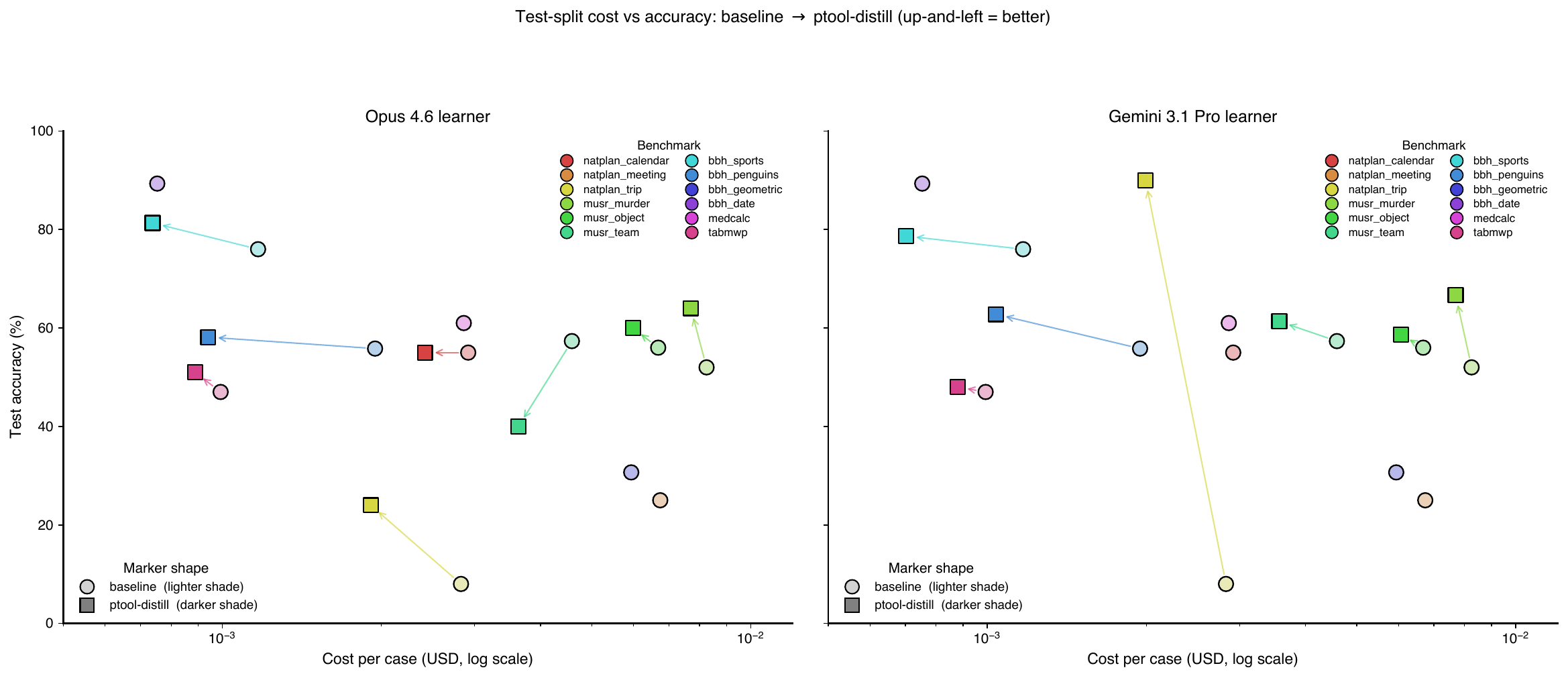}
\caption{\small Left, test split cost vs accuracy of \emph{ptool-distill} under the Claude Opus 4.6~\cite{anthropic2026opus46} learner; right, under Gemini 3.1 Pro Preview. The lighter circle is the original hand-written workflow, the darker square is the same workflow after all ptools (excluding the root interfaces)have been processed by code distillation Arrows pointing up-and-left are  lower cost and higher accuracy. Benchmarks means do ptools distillations were accepted.}
\label{fig:codedistill-ptool-test}

\includegraphics[width=0.4\textwidth]{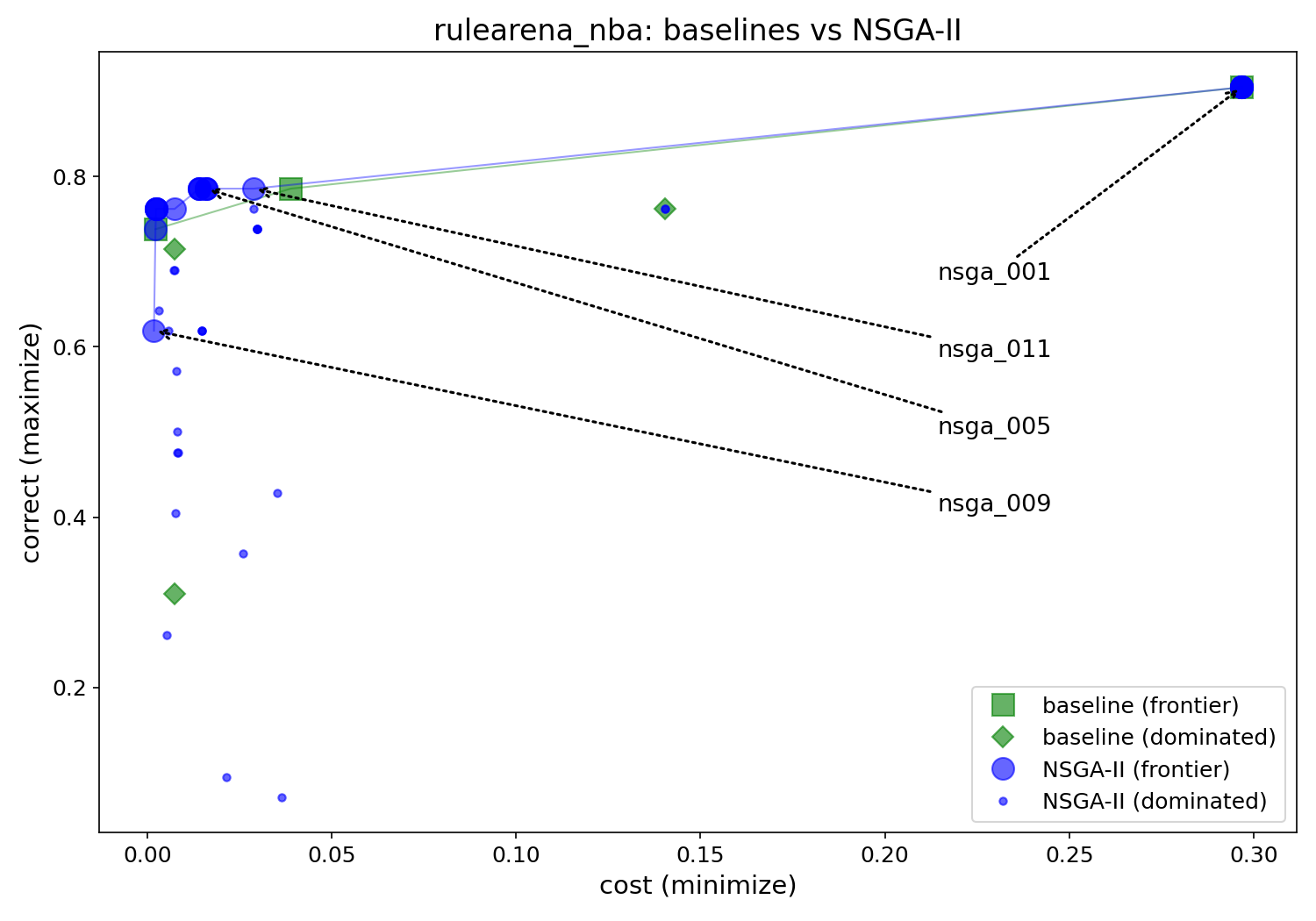}
\includegraphics[width=0.4\textwidth]{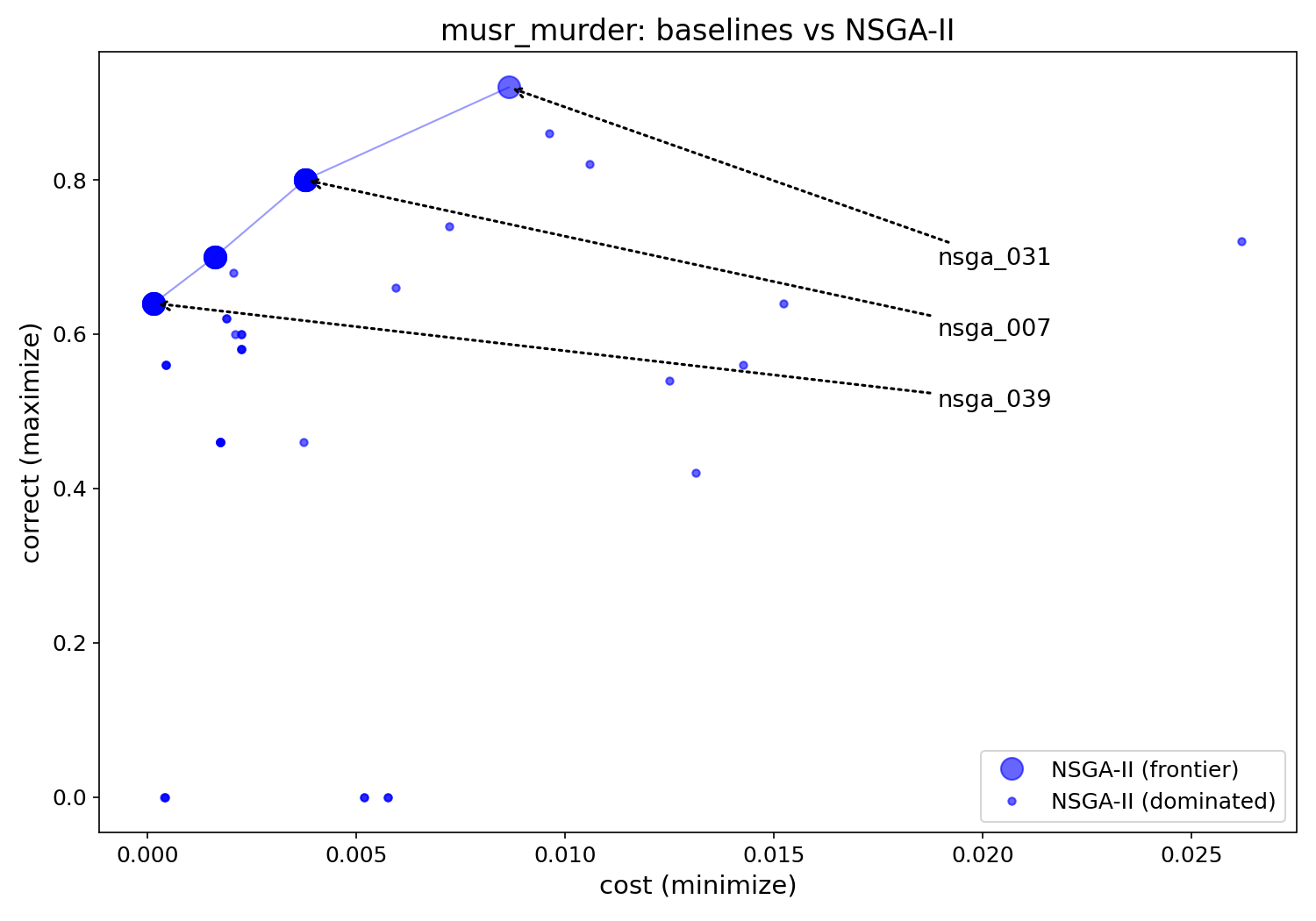}
\caption{\small Pareto-optimal configurations found for two benchmarks by the NSGA-II optimized. Left: For NBA, green points are engineered workflows with a single LLM; config 0001 switched between DeepSeek-V3 and Gemini 2.5 Flash-Lite~\cite{gemini25}, using induced ptools and ReAct; configs 011, 005 are the engineered baseline with ptools routed to a gpt-oss-20b~\cite{gptoss} model; config 009 is the same but only using gpt-oss-20b. Right: Pareto curve for a second representative benchmark, MuSR Murder.  Two of the Pareto-optimal configurations are from the orchestration learner of \S\ref{sec:learn-workflow}, one bound to Gemini 2.5 Flash-Lite and one to gpt-oss-120b, with similar output quality for $\approx$ 1/7 the cost. } \label{fig:pareto}
\end{figure*}

\textbf{Many pseudo-tools can be replaced with Python tools}. We applied the code distillation learner on handed-coded fixed workflows for 13 benchmarks. Under Opus, $12$ ptools across $7$ benchmarks (more than half of those tested) have ptools that are substantively modified (after all gating checks)--i.e., 7 ptools are algorithmic enough to express in code. Figure~\ref{fig:codedistill-ptool-test} plots results. 

\textbf{Modularity makes bi-objective optimization possible.}
As described in \S\ref{sec:bi-objective}, we used the NSGA-II evolutionary search method to search the space of possible configurations produced by the various learners described above. Two representative results are shown in Figure~\ref{fig:pareto}, where we show some ``frontier configurations'' that optimally tradeoff cost and output quality.

\section{Limitations and broader impacts} \label{sec:limitations}

The goal of making agentic systems more efficient and interpretable is of primary importance in AI, and we believe that this work makes some steps toward this: because static workflows are much easier to understand, and predict the behavior of, than dynamic ones, extending their capabilities is desirable.  As noted above, however, some of our methods can subvert the goal of interpretability, e.g., by generating Python "workflows" that bypass the toolkit that a designer intends to be used. Auditing learned workflows and learned toolkits for undesirable behavior is left as a topic for further research.

\section{Conclusion}
Principled design of practical agentic systems requires modularity. We implement a framework based on "pseudo-tools"—LLM calls with a restricted context—and show that hand-engineered pseudo-toolkits and fixed workflows outperform dynamic ReAct-style systems. We further show that both toolkits and workflows can be learned, that components can be automatically converted into hard-coded tools, and that learner and backend LLM choices can be configured via discrete search to select cheap, effective systems.

\subsection*{Acknowledgements}

The authors thank Google and Theta Labs (\url{https://www.thetalabs.org/}) for their generous support with compute resources.

\newpage

\bibliographystyle{plainnat}
\bibliography{references}

@inproceedings{zhuge2024gptswarm,
  title={Gptswarm: Language agents as optimizable graphs},
  author={Zhuge, Mingchen and Wang, Wenyi and Kirsch, Louis and Faccio, Francesco and Khizbullin, Dmitrii and Schmidhuber, J{\"u}rgen},
  booktitle={Forty-first International Conference on Machine Learning},
  year={2024}
}

@article{choure2025agentic,
  title={Agentic AI for Emergency Response and Comparative Analysis of SmolAgents, LangGraph, AutoGen, Agno AGI and CrewAI for Crisis Solution},
  author={Choure, Purvi and Prajapat, Shaligram},
  journal={Authorea Preprints},
  year={2025},
  publisher={Authorea}
}

@article{chen2022program,
  title={Program of thoughts prompting: Disentangling computation from reasoning for numerical reasoning tasks},
  author={Chen, Wenhu and Ma, Xueguang and Wang, Xinyi and Cohen, William W},
  journal={arXiv preprint arXiv:2211.12588},
  year={2022}
}

@misc{fang2025comprehensivesurveyselfevolvingai,
      title={A Comprehensive Survey of Self-Evolving AI Agents: A New Paradigm Bridging Foundation Models and Lifelong Agentic Systems}, 
      author={Jinyuan Fang and Yanwen Peng and Xi Zhang and Yingxu Wang and Xinhao Yi and Guibin Zhang and Yi Xu and Bin Wu and Siwei Liu and Zihao Li and Zhaochun Ren and Nikos Aletras and Xi Wang and Han Zhou and Zaiqiao Meng},
      year={2025},
      eprint={2508.07407},
      archivePrefix={arXiv},
      primaryClass={cs.AI},
      url={https://arxiv.org/abs/2508.07407}, 
}

@article{xiao2025designbench,
  title={Designbench: A comprehensive benchmark for mllm-based front-end code generation},
  author={Xiao, Jingyu and Wang, Ming and Lam, Man Ho and Wan, Yuxuan and Liu, Junliang and Huo, Yintong and Lyu, Michael R},
  journal={arXiv preprint arXiv:2506.06251},
  year={2025}
}

@misc{zhang2026darwingodelmachineopenended,
      title={Darwin Godel Machine: Open-Ended Evolution of Self-Improving Agents}, 
      author={Jenny Zhang and Shengran Hu and Cong Lu and Robert Lange and Jeff Clune},
      year={2026},
      eprint={2505.22954},
      archivePrefix={arXiv},
      primaryClass={cs.AI},
      url={https://arxiv.org/abs/2505.22954}, 
}

@misc{pan2026measuringagentsproduction,
      title={Measuring Agents in Production}, 
      author={Melissa Z. Pan and Negar Arabzadeh and Riccardo Cogo and Yuxuan Zhu and Alexander Xiong and Lakshya A Agrawal and Huanzhi Mao and Emma Shen and Sid Pallerla and Liana Patel and Shu Liu and Tianneng Shi and Xiaoyuan Liu and Jared Quincy Davis and Emmanuele Lacavalla and Alessandro Basile and Shuyi Yang and Paul Castro and Daniel Kang and Joseph E. Gonzalez and Koushik Sen and Dawn Song and Ion Stoica and Matei Zaharia and Marquita Ellis},
      year={2026},
      eprint={2512.04123},
      archivePrefix={arXiv},
      primaryClass={cs.CY},
      url={https://arxiv.org/abs/2512.04123}, 
}

@article{yao2023react,
  title={{ReAct}: Synergizing Reasoning and Acting in Language Models},
  author={Yao, Shunyu and Zhao, Jeffrey and Yu, Dian and Du, Nan and Shafran, Izhak and Narasimhan, Karthik and Cao, Yuan},
  journal={International Conference on Learning Representations (ICLR)},
  year={2023}
}

@article{gptswarm2024,
  title={{GPTSwarm}: Language Agents as Optimizable Graphs},
  author={Zhuge, Mingchen and Wang, Wenyi and Kirsch, Louis and Faccio, Francesco and Khizbullin, Dmitrii and Schmidhuber, J{\"u}rgen},
  journal={International Conference on Machine Learning (ICML)},
  year={2024}
}

@inproceedings{zhang2025aflow,
  title={Aflow: Automating agentic workflow generation},
  author={Zhang, Jiayi and Xiang, Jinyu and Yu, Zhaoyang and Teng, Fengwei and Chen, Xionghui and Chen, Jiaqi and Zhuge, Mingchen and Cheng, Xin and Hong, Sirui and Wang, Jinlin and others},
  booktitle={International Conference on Learning Representations},
  volume={2025},
  pages={34040--34077},
  year={2025}
}

@article{hu2024adas,
  title={Automated Design of Agentic Systems},
  author={Hu, Shengran and Lu, Cong and Clune, Jeff},
  journal={arXiv preprint arXiv:2408.08435},
  year={2024}
}

@article{khattab2023dspy,
  title={{DSPy}: Compiling Declarative Language Model Calls into Self-Improving Pipelines},
  author={Khattab, Omar and Singhvi, Arnav and Maheshwari, Paridhi and Zhang, Zhiyuan and Santhanam, Keshav and Vardhamanan, Sri and Haq, Saiful and Sharma, Ashutosh and Joshi, Thomas T. and Mober, Hanna and Shah, Pawan Kumar and Edalati, Neel and Lee, Caleb and Shin, Richard and Potts, Christopher and Zaharia, Matei},
  journal={International Conference on Learning Representations (ICLR)},
  year={2024}
}

@article{yuksekgonul2025optimizing,
  title={Optimizing generative AI by backpropagating language model feedback},
  author={Yuksekgonul, Mert and Bianchi, Federico and Boen, Joseph and Liu, Sheng and Lu, Pan and Huang, Zhi and Guestrin, Carlos and Zou, James},
  journal={Nature},
  volume={639},
  number={8055},
  pages={609--616},
  year={2025},
  publisher={Nature Publishing Group UK London}
}

@inproceedings{agrawal2026gepa,
  title={Gepa: Reflective prompt evolution can outperform reinforcement learning},
  author={Agrawal, Lakshya A and Tan, Shangyin and Soylu, Dilara and Ziems, Noah and Khare, Rishi and Opsahl-Ong, Krista and Singhvi, Arnav and Shandilya, Herumb and Ryan, Michael J and Jiang, Meng and others},
  booktitle={International Conference on Learning Representations},
  volume={2026},
  pages={8479--8565},
  year={2026}
}

@article{suzgun2022challenging,
  title={Challenging {BIG}-Bench Tasks and Whether Chain-of-Thought Can Solve Them},
  author={Suzgun, Mirac and Scales, Nathan and Sch{\"a}rli, Nathanael and Gehrmann, Sebastian and Tay, Yi and Chung, Hyung Won and Chowdhery, Aakanksha and Le, Quoc and Chi, Ed and Zhou, Denny and Wei, Jason},
  journal={Findings of the Association for Computational Linguistics (ACL Findings)},
  year={2023}
}

@article{lu2022dynamic,
  title={Dynamic Prompt Learning via Policy Gradient for Semi-structured Mathematical Reasoning},
  author={Lu, Pan and Qiu, Liang and Chang, Kai-Wei and Wu, Ying Nian and Zhu, Song-Chun and Rajpurohit, Tanmay and Clark, Peter and Kalyan, Ashwin},
  journal={International Conference on Learning Representations (ICLR)},
  year={2023}
}

@article{chen2021finqa,
  title={{F}in{QA}: A Dataset of Numerical Reasoning over Financial Data},
  author={Chen, Zhiyu and Chen, Wenhu and Smiley, Charese and Shah, Sameena and Borova, Iana and Langdon, Dylan and Moussa, Reema and Beane, Matt and Huang, Ting-Hao and Routledge, Bryan and Wang, William Yang},
  journal={Conference on Empirical Methods in Natural Language Processing (EMNLP)},
  year={2021}
}

@article{zhou2025rulearena,
  title={{R}ule{A}rena: A Benchmark for Rule-Guided Reasoning with {LLM}s in Real-World Scenarios},
  author={Zhou, Ruiwen and Hua, Wenyue and Pan, Liangming and Cheng, Sitao and Wu, Xiaobao and Yu, En and Wang, William Yang},
  journal={Annual Meeting of the Association for Computational Linguistics (ACL)},
  year={2025}
}

@article{sprague2024musr,
  title={{MuSR}: Testing the Limits of Chain-of-thought with Multistep Soft Reasoning},
  author={Sprague, Zayne and Ye, Xi and Bostrom, Kaj and Chaudhuri, Swarat and Durrett, Greg},
  journal={International Conference on Learning Representations (ICLR)},
  year={2024}
}

@article{zheng2024natural,
  title={{NATURAL PLAN}: Benchmarking {LLM}s on Natural Language Planning},
  author={Zheng, Huaixiu Steven and Mishra, Swaroop and Zhang, Hugh and Chen, Xinyun and Chen, Minmin and Nova, Azade and Hou, Le and Cheng, Heng-Tze and Le, Quoc V. and Chi, Ed H. and Zhou, Denny},
  journal={arXiv preprint arXiv:2406.04520},
  year={2024}
}

@article{khandekar2024medcalcbench,
  title={{MedCalc-Bench}: Evaluating Large Language Models for Medical Calculations},
  author={Khandekar, Nikhil and Jin, Qiao and Xiong, Guangzhi and Dunn, Soren and Applebaum, Serina S. and Anwar, Zain and Sarfo-Gyamfi, Maame and Safranek, Conrad W. and Anwar, Abid A. and Zhang, Andrew and Gilson, Aidan and Singer, Maxwell B. and Dave, Amisha and Taylor, Andrew and Zhang, Aidong and Chen, Qingyu and Lu, Zhiyong},
  journal={Advances in Neural Information Processing Systems (NeurIPS) Datasets and Benchmarks Track},
  year={2024}
}

@article{tang2024medagentbench,
  title={{MedAgentBench}: A Realistic Virtual {EHR} Environment to Benchmark Medical {LLM} Agents},
  author={Jiang, Yixing and Black, Kameron C. and Geng, Gloria and Park, Danny and Zou, James and Ng, Andrew Y. and Chen, Jonathan H.},
  journal={NEJM AI},
  year={2025},
  note={arXiv:2501.14654}
}

@article{yao2024tau,
  title={{$\tau$-bench}: A Benchmark for Tool-Agent-User Interaction in Real-World Domains},
  author={Yao, Shunyu and Shinn, Noah and Razavi, Pedram and Narasimhan, Karthik},
  journal={arXiv preprint arXiv:2406.12045},
  year={2024}
}

@article{wang2024awm,
  title={Agent Workflow Memory},
  author={Wang, Zora Zhiruo and Mao, Jiayuan and Fried, Daniel and Neubig, Graham},
  journal={International Conference on Machine Learning (ICML)},
  year={2025}
}

@article{wang2025asi,
  title={Inducing Programmatic Skills for Agentic Tasks},
  author={Wang, Zora Zhiruo and Gandhi, Apurva and Neubig, Graham and Fried, Daniel},
  journal={Conference on Language Modeling (COLM)},
  year={2025}
}

@article{zhao2024expel,
  title={{ExpeL}: {LLM} Agents Are Experiential Learners},
  author={Zhao, Andrew and Huang, Daniel and Xu, Quentin and Lin, Matthieu and Liu, Yong-Jin and Huang, Gao},
  journal={AAAI Conference on Artificial Intelligence (AAAI)},
  year={2024}
}

@misc{deepseekv3,
  title={{DeepSeek-V3} Technical Report},
  author={{DeepSeek-AI}},
  year={2024},
  eprint={2412.19437},
  archivePrefix={arXiv},
  primaryClass={cs.CL},
  url={https://arxiv.org/abs/2412.19437}
}

@misc{gemini25,
  title={{Gemini 2.5}: Pushing the Frontier with Advanced Reasoning, Multimodality, Long Context, and Next Generation Agentic Capabilities},
  author={Comanici, Gheorghe and Bieber, Eric and Schaekermann, Mike and Pasupat, Ice and others},
  year={2025},
  eprint={2507.06261},
  archivePrefix={arXiv},
  primaryClass={cs.CL},
  url={https://arxiv.org/abs/2507.06261}
}

@misc{gemini31pro,
  title={{Gemini 3.1 Pro} Model Card},
  author={{Google DeepMind}},
  year={2026},
  howpublished={\url{https://deepmind.google/models/model-cards/gemini-3-1-pro/}},
  note={Accessed May 2026}
}

@misc{anthropic2026opus46,
  title={{Claude Opus 4.6} System Card},
  author={{Anthropic}},
  year={2026},
  month={February},
  howpublished={\url{https://www.anthropic.com/system-cards}},
  note={Accessed May 2026}
}

@misc{gptoss,
  title={{gpt-oss-120b} \& {gpt-oss-20b} Model Card},
  author={{OpenAI}},
  year={2025},
  eprint={2508.10925},
  archivePrefix={arXiv},
  primaryClass={cs.CL},
  url={https://arxiv.org/abs/2508.10925}
}

@article{JMLR:v13:fortin12a,
  author  = {F{\'e}lix-Antoine Fortin and Fran{\c{c}}ois-Michel De Rainville and Marc-Andr{\'e} Gardner and Marc Parizeau and Christian Gagn{\'e}},
  title   = {{DEAP}: Evolutionary Algorithms Made Easy},
  journal = {Journal of Machine Learning Research},
  year    = {2012},
  volume  = {13},
  number  = {70},
  pages   = {2171--2175},
  url     = {http://jmlr.org/papers/v13/fortin12a.html}
}

@article{nsga2,
  author={Deb, Kalyanmoy and Pratap, Amrit and Agarwal, Sameer and Meyarivan, T.},
  title={A Fast and Elitist Multiobjective Genetic Algorithm: {NSGA-II}},
  journal={IEEE Transactions on Evolutionary Computation},
  volume={6},
  number={2},
  pages={182--197},
  year={2002}
}

@misc{litellm,
  title  = {{LiteLLM}: Python {SDK} and Proxy Server to Call 100+ {LLM} {API}s in {OpenAI} Format},
  author = {{BerriAI}},
  year   = {2023},
  howpublished = {\url{https://github.com/BerriAI/litellm}}
}

@misc{pydanticai,
  title  = {{PydanticAI}: Agent Framework, the {Pydantic} Way},
  author = {{Pydantic Services Inc.}},
  year   = {2024},
  howpublished = {\url{https://github.com/pydantic/pydantic-ai}}
}

@misc{smolagents,
  title  = {{smolagents}: a smol library to build great agentic systems},
  author = {Roucher, Aymeric and Villanova del Moral, Albert and Wolf, Thomas and von Werra, Leandro and Kaunism{\"a}ki, Erik},
  year   = {2025},
  howpublished = {\url{https://github.com/huggingface/smolagents}}
}

@misc{gemini31flashlite,
  title  = {{Gemini 3.1 Flash-Lite} Model Card},
  author = {{Google DeepMind}},
  year   = {2026},
  howpublished = {\url{https://deepmind.google/models/model-cards/gemini-3-1-flash-lite/}},
  note   = {Accessed May 2026}
}

\newpage
\appendix

\section{Description of benchmarks}

\subsection{Benchmarks used} \label{app:benchmarks}

\begin{table}[ht]
\centering
\footnotesize
\setlength{\tabcolsep}{4pt}
\caption{Benchmarks used in this work. ``Tools'' counts Python or deterministic interfaces in the hand-coded workflow; ``Ptools'' counts LLM-backed pseudo-tool interfaces.}
\label{tab:benchmarks}
\begin{tabular}{lp{5cm}ccrrrl}
\toprule
Benchmark & Description & Tools & Ptools & Train & Valid & Test & Reference \\
\midrule
BBH / date understanding      & Determine a calendar date from prose temporal constraints              & 0  & 6 &     75 &     75 &     100 & \cite{suzgun2022challenging} \\
BBH / geometric shapes        & Identify a geometric figure from an ASCII art description              & 0  & 7 &     75 &     75 &     100 & \cite{suzgun2022challenging} \\
BBH / penguins in a table     & Answer attribute lookup questions over a formatted table               & 0  & 4 &     43 &     43 &      60 & \cite{suzgun2022challenging} \\
BBH / sports understanding    & Determine whether a sports-action sentence is plausible                & 0  & 3 &     75 &     75 &      75 & \cite{suzgun2022challenging} \\
\midrule
DesignBench / vanilla         & Generate HTML/CSS code matching a reference screenshot                 & 0  & 1 &     -- &     -- &     120 & \cite{xiao2025designbench} \\
DesignBench / vue             & Generate a Vue.js component matching a reference screenshot            & 0  & 1 &     -- &     -- &     118 & \cite{xiao2025designbench} \\
DesignBench / angular         & Generate an Angular component matching a reference screenshot          & 0  & 1 &     -- &     -- &      83 & \cite{xiao2025designbench} \\
\midrule
FinQA                         & Numerical reasoning over financial tables and free text                & 4  & 2 &    100 &    100 &     300 & \cite{chen2021finqa} \\
\midrule
MedAgentBench                 & Medical EHR tasks via FHIR API (lab orders, vitals, prescriptions)     & 2  & 10 &     -- &     -- &     300 & \cite{tang2024medagentbench} \\
\midrule
MedCalc / equation            & Compute a medical score given an explicit formula                      & 1  & 4 &     63 &     63 &     660 & \cite{khandekar2024medcalcbench} \\
MedCalc / rule                & Compute a medical score given a prose eligibility rule                 & 1  & 5 &     34 &     34 &     380 & \cite{khandekar2024medcalcbench} \\
\midrule
MuSR / murder mystery         & Identify a murderer from alibi and motive evidence                     & 0  & 4 &     75 &     75 &     100 & \cite{sprague2024musr} \\
MuSR / object placements      & Track object locations across a sequence of moves                      & 0  & 4 &     75 &     75 &     106 & \cite{sprague2024musr} \\
MuSR / team allocation        & Assign agents to teams under compatibility constraints                 & 0  & 3 &     75 &     75 &     100 & \cite{sprague2024musr} \\
\midrule
Natural Plan / calendar       & Schedule appointments without time conflicts                           & 0  & 3 &    100 &    100 &     100 & \cite{zheng2024natural} \\
Natural Plan / meeting        & Find a meeting slot satisfying all participants' availability          & 0  & 3 &    100 &    100 &     100 & \cite{zheng2024natural} \\
Natural Plan / trip           & Plan a multi-city trip within time and budget constraints              & 0  & 3 &    100 &    100 &     100 & \cite{zheng2024natural} \\
\midrule
RuleArena / airline           & Compute baggage fees and ticket total from American Airlines fee rules & 1  & 1 &     50 &     50 &     100 & \cite{zhou2025rulearena} \\
RuleArena / tax               & Compute US federal income tax owed from filled IRS form inputs         & 1  & 1 &     50 &     50 &     100 & \cite{zhou2025rulearena} \\
RuleArena / nba               & Determine whether proposed team operations violate NBA CBA rules       & 0  & 1 &     50 &     42 &      46 & \cite{zhou2025rulearena} \\
\midrule
TabMWP                        & Arithmetic word problems requiring lookup in a structured table        & 0  & 2 &    100 &    100 &     100 & \cite{lu2022dynamic} \\
\midrule
Tau Bench / retail            & Multi-turn retail customer service with order and account operations   & 16 & 7 &     -- &     60 &      54 & \cite{yao2024tau} \\
\bottomrule
\end{tabular}
\end{table}

\newpage




\section{Additional experimental results} \label{app:hero}


\subsection{Other evaluated strategies}

In addition to the results of Table~\ref{tab:hero}, we compared the static workflow engineered with two other agent strategies.  

One was a traditional zero-shot prompt.  This allows more flexibility in prompt tuning, which was often successful.

Recall that one supported implementation factory generates a (possibly unique) Python function for each call to an interface, and executes that function in a sandbox.  When an empty toolkit is used, this implements 
program of thoughts \cite{chen2022program} (PoT). Replacing the empty toolkit with the same kit of tools and pseudo-tools used by the static workflow and ReAct does an intermediate amount of planning: a new workflow is constructed for each problem instance, but the workflow is constructed all at once, rather than incrementally as tool outputs are seen, as in ReAct.  Tables~\ref{tab:hero-correctness} and \ref{tab:hero-cost} show this dynamic workflow variant is intermediate in average performance between ReAct and the zero-shot default prompt: however, there are cases where it behaves quite differently to ReAct.

\begin{table*}[ht]
\centering
\begin{small}
\caption{Result quality (generally accuracy) across multiple benchmarks for different root interface bindings, using DeepSeek V3-1.  The static workflow is hand-engineered on a dev set.  The dynamic workflows use the same pseudo-tools and tools: ReAct replans after each tool or ptool output is observed, and PoT constructs a complete tool-calling plan for each task instance before running and tools.  The default implementation zero-shot baseline uses one non-agentic LLM call, without demonstrations, and the custom prompt sometimes uses a second LLM call to extract a final answer from the first call.} 
\label{tab:hero-correctness}
\begin{tabular}{lcccccl}
\toprule
Task & \makecell{Static\\Workflow} & \makecell{Dynamic\\Workflow\\(ReAct)} & \makecell{Dynamic\\Workflow\\(PoT)} & \makecell{Zero-shot\\(Default\\Imp.)} & \makecell{Zero-shot\\(Traditional\\Prompt)} & \makecell{Zero-shot\\(Traditional\\Prompt\\w/ Thinking)} \\
\midrule
BBH Date Understanding & $0.84$ & $\mathbf{0.93}$ & $0.82$ & $0.52$ & $0.88$ & $0.88$ \\
BBH Geometric Shapes & $0.30$ & $0.35$ & $0.43$ & $0.53$ & $\mathbf{0.63}$ & $\mathbf{0.63}$ \\
BBH Penguins in a Table & $0.63$ & $0.72$ & $0.25$ & $0.93$ & $\mathbf{0.98}$ & $\mathbf{0.98}$ \\
BBH Sports Understanding & $0.87$ & $\mathbf{0.88}$ & $0.71$ & $0.65$ & $0.84$ & $0.84$ \\
FinQA & $\mathbf{0.75}$ & $0.32$ & $0.18$ & $0.62$ & $0.57$ & $0.57$ \\
MedAgentBench & $\mathbf{0.87}$ & $0.69$ & $0.49$ & $0.00$ & $0.00$ & $0.00$ \\
MedCalc Formulas & $0.81$ & $0.82$ & $0.75$ & $0.60$ & $0.33$ & $\mathbf{0.86}$$^*$ \\
MedCalc Rules & $0.50$ & $0.47$ & $0.43$ & $0.45$ & $0.26$ & $\mathbf{0.62}$$^*$ \\
MUSR Murder & $\mathbf{0.68}$ & $0.62$ & $0.65$ & $0.52$ & $0.37$ & $0.46$$^*$ \\
MUSR Objects & $0.58$ & $0.36$ & $\mathbf{0.67}$ & $0.35$ & $0.31$ & $0.37$$^*$ \\
MUSR Teams & $\mathbf{0.61}$ & $0.49$ & $0.52$ & $0.59$ & $0.60$ & $0.59$$^*$ \\
NaturalPlan Calendar & $0.62$ & $0.46$ & $0.56$ & $0.54$ & $\mathbf{0.70}$ & $\mathbf{0.70}$ \\
NaturalPlan Meeting & $\mathbf{0.37}$ & $0.23$ & $0.28$ & $0.23$ & $0.32$ & $0.32$ \\
NaturalPlan Trip & $\mathbf{0.21}$ & $0.16$ & $\mathbf{0.21}$ & $0.17$ & $0.01$ & $0.01$ \\
Rulearena Airlines & $\mathbf{0.98}$ & $0.89$ & $0.84$ & $0.41$ & $0.43$ & $0.46$$^*$ \\
Rulearena Tax & $0.50$ & $0.15$ & $0.43$ & $0.11$ & $0.47$ & $\mathbf{0.55}$$^*$ \\
Rulearena NBA & $0.61$ & $0.59$ & $0.02$ & $0.67$ & $\mathbf{0.78}$ & $0.72$$^*$ \\
Tabular Math WP & $0.90$ & $\mathbf{0.97}$ & $0.94$ & $0.87$ & $0.96$ & $0.96$ \\
$\tau$ Bench Retail & $0.52$ & $\mathbf{0.56}$ & $0.15$ & $0.11$ & $0.11$ & $0.11$ \\
\midrule
\textrm{Average} & $\mathbf{0.64}$ & $0.56$ & $0.49$ & $0.47$ & $0.50$ & $0.56$ \\
\bottomrule
\end{tabular}
\end{small}

\end{table*}

\begin{table*}[ht]
\centering
\caption{Cost (per 100 examples) for different root interface bindings, using DeepSeek V3-1.} \label{tab:hero-cost}
\begin{small}
\begin{tabular}{lcccccl}
\toprule
Task & \makecell{Static\\Workflow} & \makecell{Dynamic\\Workflow\\(ReAct)} & \makecell{Dynamic\\Workflow\\(PoT)} & \makecell{Zero-shot\\(Default\\Imp.)} & \makecell{Zero-shot\\(Traditional\\Prompt)} & \makecell{Zero-shot\\(Traditional\\Prompt\\w/ Thinking)} \\
\midrule
BBH Date Understanding & $0.28$ & $0.99$ & $0.40$ & $0.10$ & $\mathbf{0.08}$ & $\mathbf{0.08}$ \\
BBH Geometric Shapes & $0.63$ & $2.40$ & $0.64$ & $\mathbf{0.07}$ & $\mathbf{0.07}$ & $\mathbf{0.07}$ \\
BBH Penguins in a Table & $0.21$ & $0.54$ & $0.31$ & $\mathbf{0.05}$ & $\mathbf{0.05}$ & $\mathbf{0.05}$ \\
BBH Sports Understanding & $0.13$ & $0.29$ & $0.21$ & $\mathbf{0.02}$ & $0.03$ & $0.03$ \\
DesignBench Vanilla & -- & $0.29$ & -- & -- & $\mathbf{0.09}$ & $\mathbf{0.09}$ \\
FinQA & $0.12$ & $0.95$ & $0.35$ & $0.14$ & $\mathbf{0.10}$ & $\mathbf{0.10}$ \\
MedAgentBench & $0.64$ & $0.65$ & $0.82$ & $0.22$ & $\mathbf{0.02}$ & $\mathbf{0.02}$ \\
MedCalc Formulas & $0.15$ & $1.20$ & $0.47$ & $0.19$ & $\mathbf{0.12}$ & $0.20$$^*$ \\
MedCalc Rules & $0.27$ & $1.29$ & $0.54$ & $0.21$ & $\mathbf{0.13}$ & $0.24$$^*$ \\
MUSR Murder & $0.47$ & $1.34$ & $0.80$ & $\mathbf{0.11}$ & $0.14$ & $0.16$$^*$ \\
MUSR Objects & $0.34$ & $0.99$ & $0.39$ & $\mathbf{0.10}$ & $0.15$ & $0.13$$^*$ \\
MUSR Teams & $0.28$ & $1.38$ & $0.35$ & $\mathbf{0.14}$ & $0.28$ & $0.21$$^*$ \\
NaturalPlan Calendar & $0.46$ & $0.79$ & $0.63$ & $\mathbf{0.09}$ & $0.26$ & $0.26$ \\
NaturalPlan Meeting & $0.57$ & $0.91$ & $0.79$ & $\mathbf{0.11}$ & $0.27$ & $0.27$ \\
NaturalPlan Trip & $0.37$ & $1.19$ & $0.52$ & $\mathbf{0.08}$ & $0.35$ & $0.35$ \\
Rulearena Airlines & $\mathbf{0.97}$ & $2.40$ & $1.14$ & $1.12$ & $1.30$ & $1.13$$^*$ \\
Rulearena Tax & $1.01$ & $3.49$ & $1.67$ & $\mathbf{0.68}$ & $1.01$ & $0.98$$^*$ \\
Rulearena NBA & $1.49$ & $2.95$ & $1.85$ & $\mathbf{1.42}$ & $1.59$ & $1.62$$^*$ \\
Tabular Math WP & $0.07$ & $0.69$ & $0.10$ & $0.04$ & $\mathbf{0.03}$ & $\mathbf{0.03}$ \\
$\tau$ Bench Retail & $4.75$ & $7.03$ & $5.14$ & $3.50$ & $\mathbf{3.15}$ & $\mathbf{3.15}$ \\
\midrule
\textrm{Average} & $0.70$ & $1.59$ & $0.90$ & $\mathbf{0.44}$ & $0.46$ & $0.46$ \\
\bottomrule
\end{tabular}
\end{small}

\end{table*}

We also considered a baseline that uses a single user-constructed and tuned prompt. This baseline is also surprisingly strong for many tasks; the tables above give the full details of the results from Figure~\ref{fig:workflow}.

\subsection{Pseudo-tool induction experiments} \label{app:ptool-ind-exp}

For each subtask we evaluated two action spaces under the same
ReAct framework (\texttt{react\_pydantic}, a clean ReAct preamble
over a pydantic-ai loop), the same backbone model
(DeepSeek-V3), and the same test split (seed~42):
\emph{engineered}, the hand-written multi-step pipeline, and
\emph{induced}, a $K{=}5$ ptool library induced from the
engineered agent's correct training rollouts using the pipeline
of \S\ref{sec:learn-pseudo-toolkit}.\footnote{%
NaturalPlan accuracy is reported under a Gemini-2.5-flash LLM judge,
since the strict format-matching evaluator zero-rates almost all
induced plans; engineered runs are scored under the same judge for
apples-to-apples.  The MedCalc induced module was synthesized by
Gemini 3.1 Flash-Lite~\cite{gemini31flashlite} (state-aware,
correct-only, $K{=}5$ cap), one LLM-drift point relative to the other
tasks where induction used DeepSeek-V3.}

\begin{table}[ht]
\centering
\small
\caption{Cross-benchmark induced-ptool evaluation.  Both columns use
the same ReAct framework (\texttt{react\_pydantic}), backbone model
(DeepSeek-V3), and test split (seed~42); the only difference is the
action space (hand-engineered multi-step pipeline vs.\ a $K{=}5$
induced ptool library, correct-only filter).  Bold marks the higher
of each row.  NaturalPlan rows are scored under a Gemini-2.5-flash
LLM judge for apples-to-apples; others use strict exact-match.}
\label{tab:induction-cross-benchmark}
\begin{tabular}{lccc}
\toprule
Subtask & Engineered (col~3) & Induced (col~4) & $\Delta$ \\
\midrule
MuSR murder              & 57.0 & \textbf{75.0} & +18.0 \\
MuSR object placements   & 30.2 & \textbf{68.9} & +38.7 \\
MuSR team allocation     & 44.0 & \textbf{68.0} & +24.0 \\
NaturalPlan meeting      & 10.0 & \textbf{35.0} & +25.0 \\
NaturalPlan trip         & \textbf{6.0} & 1.0 & $-5.0$ \\
RuleArena NBA            & 65.2 & \textbf{71.7} & +6.5 \\
MedCalc                  & 44.0 & \textbf{54.0} & +10.0 \\
\bottomrule
\end{tabular}

\end{table}

The pattern matches the bad-tools framing.  On MuSR, where the
engineered toolkit consists of reasoning-only steps with no external
grounding, induction wins by 18--39\,pp across all three subtasks,
peaking at $+38.7$\,pp on object-placements where the engineered
3-step pipeline is the most abstract.  On NaturalPlan meeting, where
the hand-designed parse-order-build pipeline collapses under the
\texttt{react\_pydantic} framing (10\%), induction sustains $35\%$ and
wins by $+25$\,pp---the engineered pipeline's value is recoverable
from rollouts but not robust to a clean ReAct prompt.  On NBA, where
the engineered toolkit is constrained to a single one-shot extraction
tool because the domain has no calculator, induction still wins but
by only $+6.5$\,pp---a single typed extraction is most of what the
task needs, so additional structure pays small dividends.  On MedCalc,
induction wins by $+10$\,pp; the gain is moderate because the
hand-written calculator-identification pipeline is genuinely
informative on the formula-driven subset of cases, even under
\texttt{react\_pydantic}.

NaturalPlan trip is the lone reversal ($-5$\,pp), but both methods
saturate near zero ($1$--$6\%$): the eval is dominated by harder
multi-day, multi-constraint instances on which neither action space
helps the LLM.  We treat it as a non-comparison cell rather than
evidence against the framing; per-method differences below
$10\%$ on a noisy LLM-judge metric are within the regime where
neither system reliably solves the task.

The overall picture is that induction's leverage tracks the
\emph{abstractness} of the engineered toolkit: largest gains where
hand-designed tools are reasoning steps without external grounding
(MuSR, +24\,pp average), substantial gains where the engineered
pipeline collapses under a clean ReAct framing (NatPlan meeting,
+25\,pp), moderate gains where engineered tools have natural domain
structure (MedCalc, NBA), and indistinguishable in the saturated-low
regime (NatPlan trip).

\subsection{Workflow learning experiments} \label{app:workflow-learning-expt}

We tested the ptool inducer on MedCalc, where the hand-written ReAct toolkit decomposes the problem into three interfaces (\textsf{identify\_calculator}, \textsf{extract\_clinical\_values}, \textsf{compute\_calculation}). Inducing from $101$ recorded thoughts over $110$ training problems and a held-out evaluation of $12$ induction variants (axes: state injection, only-correct filtering, ptool cap), the best variant collapsed the three-step pipeline into three self-contained ptools (\textsf{calculate\_clinical\_score}, \textsf{compute\_clinical\_value}, \textsf{apply\_clinical\_score}) and improved within-tolerance accuracy from $58.2\%$ to $66.4\%$ while reducing input tokens by $60\%$ (2{,}129K to 848K). The induced ptools recover most of the gain on formula-driven categories (\emph{physical} $+15.6$pp, \emph{lab test} $+12.5$pp) but do not improve criterion-counting categories. The induction itself cost $\sim\$0.30$ over $\sim 10$ minutes of wall-clock; replacing several hand-written tools with their induced counterparts is therefore a cheap and reversible operation. Composing these induced ptools through the orchestration learner of \S\ref{sec:learn-workflow} closes the remaining gap to the hand-coded workflow on this benchmark (\S\ref{sec:learn-workflow-results}). 

We applied the orchestration learner of \S\ref{sec:learn-workflow} to $15$ benchmarks, in two modes per benchmark: \emph{existing} (start from the hand-engineered workflow) and \emph{seeded} (compose a workflow from the benchmark's (pseudo-)toolkit at iteration zero, then improve). Each run uses six iterations and Gemini~3.1~Pro as supervisor. Across the $30$ runs, eval accuracy improved over the iteration-zero baseline on $16$ runs and was preserved on the rest; on $9$ runs the supervisor consistently produced no applicable edit, mostly on RuleArena where the existing workflow is already specialized to a small number of rule paths. The tool-seeded mode is the more interesting comparison: on several compositional benchmarks the workflow induced from the toolkit alone matches or exceeds the hand-engineered workflow---NaturalPlan trip ($100\%$ vs.\ $50\%$ best eval), NaturalPlan calendar ($100\%$ vs.\ $70\%$), MuSR team allocation ($86.7\%$ vs.\ $63.3\%$), Geometric Shapes ($100\%$ vs.\ $86.4\%$), and MedCalc ($74.4\%$ vs.\ $72.0\%$). On MedCalc, where we ran a deeper six-iteration study from induced ptools, the supervisor's accepted edits mostly took the form of richer docstrings encoding formula references and scoring criteria; in particular, an iteration that added a deterministic Python expression evaluator as a ptool produced the run's best eval ($62.7\%$). These results indicate that, given a serviceable toolkit, supervised hill-climbing over workflow source can match hand-engineering on a non-trivial fraction of tasks, while making the entire workflow---and the rationale for each edit---inspectable as a code diff. The full sweep table, per-iteration accuracy curves, and the MedCalc iteration log appear in Appendix~\ref{app:orchestrator-sweep}.

\section{The orchestration learner}\label{app:orchestrator}

This appendix describes the orchestration learner of \S\ref{sec:learn-workflow} in full: its inputs, its per-iteration loop, the supervisor prompt, the accept/reject rule, the workflow-composer used to produce a seed in the \emph{seeded} mode, the artifacts written for each run, and the implementation details that affect reproducibility.

\subsection{Inputs and starting state}

A run is parameterized by:
\begin{itemize}
\item A \emph{ptools module}, i.e.\ a Python file declaring a set of \texttt{@interface} stubs together with their bindings. The entry interface is named explicitly; all other interfaces in the module are available to the supervisor as building blocks.
\item A \emph{starting workflow}, in one of two modes. In workflow-seeded mode the starting source is the benchmark's hand-coded workflow. In tool-seeded mode the starting source is a workflow synthesized at iteration zero by the workflow composer of \ref{app:orchestrator-composer} from the (pseudo-)tools available in the module.
\item A \emph{train} dataset and an \emph{eval} dataset, both stratified by a benchmark-specific label. We used $n_{\mathrm{train}}\!=\!n_{\mathrm{eval}}\!=\!110$ for MedCalc and either $70/35$ or $110/110$ for the other benchmarks.
\item A \emph{supervisor model} (we used \texttt{gemini-3.1-pro-preview}) and an iteration budget (six in all reported runs). Optional inputs include \emph{custom instructions} (free-text guidance about pipeline strengths/weaknesses, or hard rules such as ``do not modify the deterministic calculator'') and a \emph{model-choice catalog} (cheap and expensive alternatives that the supervisor may recommend in configuration overrides).
\end{itemize}

The starting source is copied to a scratch file (\texttt{<name>\_<timestamp>\_scratch.py}) and all subsequent edits target that file; the original ptools module is never modified. Iteration zero is a baseline evaluation of the starting workflow on both train and eval, recorded as a \texttt{KEPT} record.

\subsection{Per-iteration loop}

At iteration $t \ge 1$:
\begin{enumerate}
\item \textbf{Profile.} Re-run the most recent train evaluation through a profiler that aggregates per-interface call counts, per-interface average cost, and overall accuracy.
\item \textbf{Sample failures.} From the same train run, sample failed cases ($\texttt{correct}\!=\!\textsc{false}$) and format each as $\langle$input, predicted answer, expected answer, ordered ptool calls$\rangle$.
\item \textbf{Build the supervisor prompt}, which contains:
  \begin{itemize}
  \item the full source of the current scratch ptools file, verbatim;
  \item the profiling summary;
  \item the sampled failure traces;
  \item the iteration history---a concise per-iteration record of \textsc{kept}/\textsc{rolled-back}, the supervisor's stated reasoning, and the accuracy delta, so that the supervisor does not propose the same change twice;
  \item the custom instructions and the model-choice catalog if provided.
  \end{itemize}
\item \textbf{Call the supervisor.} A single LLM call returns one of: (a) a full rewritten ptools source plus optional configuration overrides (e.g.\ a per-interface model assignment); (b) a no-change signal; or (c) a syntactically invalid response. (b) and (c) increment a no-improvement counter; the run halts when this counter reaches five.
\item \textbf{Apply.} Write the proposed source over the scratch file and reload it via \texttt{exec\_ptools\_module}; re-bind every interface declared in the new source through \texttt{implement\_via\_config}; if configuration overrides were proposed, snapshot the global configuration and apply them.
\item \textbf{Re-evaluate} on train; then on eval if eval is available.
\item \textbf{Decide.} The proposal is \emph{kept} iff it strictly improves the best train accuracy seen so far, with the eval accuracy used to break ties when train is matched:
{\small
\begin{verbatim}
if new_train > best_train:                      kept = True
elif new_train == best_train and have_eval:     kept = (new_eval > best_eval)
else:                                           kept = False
\end{verbatim}}
On a rollback the scratch file is restored from the previous iteration's source, the configuration snapshot is restored, and the module is reloaded again so subsequent iterations see the previous code. The supervisor's reasoning is retained in the iteration history regardless of the decision, so it can inform later proposals.
\end{enumerate}

\paragraph{Stop conditions.} The loop ends when (i) a target accuracy is reached (rare and usually unset), (ii) the no-improvement counter reaches five, (iii) five consecutive supervisor failures or no-change responses occur, or (iv) the iteration budget is exhausted.

\paragraph{Final eval.} After the in-process loop terminates, the best-train iteration's source becomes \texttt{ptools\_evolved.py} and a final evaluation is launched in a \emph{fresh subprocess} on the eval set. The fresh process is necessary because Python's \texttt{importlib.reload} does not reset modules loaded via \texttt{spec\_from\_file\_location}; in-process iterations would otherwise produce duplicate-tool-name errors when a previously executed source defines tools with the same names. Within-iteration evals during the loop are unaffected because the previous source is replaced rather than added; the conflict surfaces only when a downstream consumer attempts to register tools whose names overlap with an earlier iteration still resident in module memory.

\subsection{Workflow composer (used in tool-seeded mode)}\label{app:orchestrator-composer}

The composer turns a (pseudo-)toolkit into an initial pipeline source.

\begin{enumerate}
\item Build a \emph{catalog} from the ptools module by iterating its declared interfaces, excluding the entry interface, and rendering each as ``signature + docstring''.
\item Issue a single LLM call (default \texttt{Gemini 3.1 Pro Preview}) with a prompt containing the task description, the catalog, and the desired entry signature; the call returns Python source for the entry function, expected to invoke each catalog interface as a typed call so that recording, caching, and cost tracking continue to work.
\item Run \texttt{ruff~--fix} on the returned source for deterministic cleanup (unused imports, formatting).
\item Optionally smoke-test the composed source on one example case; on exception, retry up to three times by feeding the error back to the model along with the prior source.
\item Bind the composed function to the entry interface; the source becomes the iteration-zero scratch file.
\end{enumerate}

The composer thus performs \emph{static} workflow induction in a single call; the orchestration learner performs \emph{iterative} repair on top of it. Because the orchestration learner uses a stronger supervisor model (Gemini~3.1~Pro) than the composer, the design separates a fast cheap structural pass from a slow expensive refinement pass.

\subsection{Per-iteration artifacts and run-level outputs}

Every iteration directory \texttt{iterations/iter\_NNN/} contains:
\begin{itemize}
\item \texttt{ptools\_before.py}, \texttt{ptools\_after.py}, and a unified diff;
\item \texttt{supervisor\_prompt.txt}, \texttt{supervisor\_response.txt};
\item \texttt{profiling\_summary.txt}, \texttt{failure\_traces.txt}, \texttt{iteration\_history.txt};
\item \texttt{outcome.txt} recording the decision (\textsc{kept}/\textsc{rolled-back}), the train and eval deltas, and the supervisor's reasoning;
\item \texttt{config\_before.yaml} and \texttt{config\_after.yaml} when configuration overrides were proposed;
\item \texttt{result\_dirs.json} pointing at the train and eval result directories produced this iteration.
\end{itemize}

The run directory itself contains a machine-readable \texttt{report.json}, a self-contained HTML dashboard with an accuracy curve and per-iteration drilldown, the best-train \texttt{ptools\_evolved.py}, an \texttt{implementation.yaml} that points at the evolved file as a learned implementation, a \texttt{run\_metadata.json}, and the fresh-process \texttt{final\_eval/} subdirectory. Because every prompt, response, and intermediate evaluation is preserved, any iteration can be replayed or audited offline; the run is reproducible up to LLM nondeterminism.

\subsection{Sweep results}\label{app:orchestrator-sweep}

We ran the learner on $15$ (benchmark, entry-interface) pairs in both modes for $30$ runs total, six iterations each. Initial T/E is the iteration-zero train/eval; Best E,T/E is the iteration that maximized eval (and its train/eval); Final T/E is the last accepted iteration. The TabMWP final-eval entries marked $0.0\%^{\dagger}$ are an instrumentation artefact: a one-time setup hook populates an in-memory table store, but the fresh-process final eval reloads the evolved module \emph{after} the hook has run, clearing the store; the per-iteration evals (which share the in-process store) are correct, and the workflow is not actually regressing.

\begin{table}[ht]
\centering
\footnotesize
\caption{Orchestration-learner sweep: 30 runs, $\le\!6$ iterations each. T/E denotes train/eval accuracy as a percentage. ``Workflow-seeded'' starts from the hand-engineered workflow; ``Tool-Seeded'' starts from a composer-generated workflow over the same toolkit.}
\label{tab:orchestrator-sweep}
\begin{tabular}{llrrr}
\toprule
Mode & Benchmark & Initial T/E & Best E iter / T / E & Final T/E \\
\midrule
wf-seed & finqa                   & 66.2 / 58.9 & 0 /\ 66.2 /\ 58.9 & 66.2 / 58.9 \\
wf-seed & medcalc                 & 79.3 / 65.9 & 4 /\ 76.7 /\ \textbf{72.0} & 79.3 / 67.1 \\
wf-seed & musr / murder           & 72.9 / 73.3 & 4 /\ 70.0 /\ \textbf{86.7} & 80.0 / 80.0 \\
wf-seed & musr / object           & 55.4 / 46.9 & 5 /\ 77.0 /\ \textbf{78.1} & 77.0 / 78.1 \\
wf-seed & musr / team             & 72.9 / 63.3 & 0 /\ 72.9 /\ 63.3 & 72.9 / 63.3 \\
wf-seed & natural\_plan / calendar & 55.7 / 70.0 & 0 /\ 55.7 /\ 70.0 & 55.7 / 70.0 \\
wf-seed & natural\_plan / meeting  & 31.4 / 33.3 & 1 /\ \textbf{100.0} /\ \textbf{100.0} & 100.0 / 100.0 \\
wf-seed & natural\_plan / trip     & 47.1 / 50.0 & 0 /\ 47.1 /\ 50.0 & 47.1 / 50.0 \\
wf-seed & rulearena / airline     & 28.6 / 16.7 & 0 /\ 28.6 /\ 16.7 & 28.6 / 16.7 \\
wf-seed & rulearena / nba         & 89.7 / 61.5 & 0 /\ 89.7 /\ 61.5 & 89.7 / 61.5 \\
wf-seed & rulearena / tax         & 0.0 / 0.0   & 0 /\ 0.0 /\ 0.0 & 0.0 / 0.0 \\
wf-seed & tabmwp                  & 54.3 / 50.0 & 0 /\ 54.3 /\ 50.0 & 54.3 / $0.0^{\dagger}$ \\
wf-seed & sports\_understanding   & 90.6 / 100.0 & 0 /\ 90.6 /\ 100.0 & 100.0 / 100.0 \\
wf-seed & geometric\_shapes       & 54.7 / 50.0 & 1 /\ 94.3 /\ \textbf{86.4} & 96.2 / 86.4 \\
wf-seed & penguins\_in\_a\_table  & 86.7 / 53.8 & 5 /\ 100.0 /\ \textbf{92.3} & 100.0 / 92.3 \\
\midrule
tool-seed   & finqa                   & 66.2 / 63.3 & 4 /\ 74.3 /\ \textbf{68.9} & 74.3 / 68.9 \\
tool-seed   & medcalc                 & 56.0 / 48.8 & 4 /\ 81.9 /\ \textbf{74.4} & 82.9 / 72.0 \\
tool-seed   & musr / murder           & 72.9 / 73.3 & 5 /\ 77.1 /\ \textbf{90.0} & 77.1 / 90.0 \\
tool-seed   & musr / object           & 55.4 / 46.9 & 5 /\ 68.9 /\ \textbf{62.5} & 68.9 / 62.5 \\
tool-seed   & musr / team             & 72.9 / 63.3 & 3 /\ 85.7 /\ \textbf{86.7} & 85.7 / 86.7 \\
tool-seed   & natural\_plan / calendar & 55.7 / 66.7 & 4 /\ 85.7 /\ \textbf{100.0} & 88.6 / 93.3 \\
tool-seed   & natural\_plan / meeting  & 31.4 / 33.3 & 4 /\ 82.9 /\ \textbf{83.3} & 82.9 / 83.3 \\
tool-seed   & natural\_plan / trip     & 47.1 / 50.0 & 4 /\ \textbf{100.0} /\ \textbf{100.0} & 100.0 / 100.0 \\
tool-seed   & rulearena / airline     & 0.0 / 0.0   & 0 /\ 0.0 /\ 0.0 & 0.0 / 0.0 \\
tool-seed   & rulearena / nba         & 89.7 / 76.9 & 0 /\ 89.7 /\ 76.9 & 89.7 / 76.9 \\
tool-seed   & rulearena / tax         & 0.0 / 0.0   & 0 /\ 0.0 /\ 0.0 & 0.0 / 0.0 \\
tool-seed   & tabmwp                  & 55.7 / 50.0 & 0 /\ 55.7 /\ 50.0 & 55.7 / $0.0^{\dagger}$ \\
tool-seed   & sports\_understanding   & 90.6 / 100.0 & 0 /\ 90.6 /\ 100.0 & 100.0 / 100.0 \\
tool-seed   & geometric\_shapes       & 52.8 / 54.5 & 5 /\ 100.0 /\ \textbf{100.0} & 100.0 / 100.0 \\
tool-seed   & penguins\_in\_a\_table  & 86.7 / 53.8 & 3 /\ 83.3 /\ \textbf{92.3} & 100.0 / 76.9 \\
\bottomrule
\end{tabular}
\end{table}

In aggregate $16$ of $30$ runs improved best-eval over the iteration-zero baseline, and $16$ retained that improvement at end of run; on $9$ of $30$ runs the supervisor produced no applicable code edit across multiple iterations, concentrated on RuleArena where the existing workflow is already specialized to a small set of rule paths. On six of the fifteen benchmarks the tool-seeded mode reached or exceeded the workflow-seeded mode (FinQA, MedCalc, MuSR team, NaturalPlan~$\{$calendar, meeting, trip$\}$, Geometric Shapes), supporting the claim that workflow induction from a sound toolkit can match or surpass hand-engineered workflows on tasks with compositional structure.

\section{Optimization details}

\subsection{Cross-benchmark frontier summary}

In our optimization experiments (Table~\ref{tab:opt-cross}) we ran the NSGA-II optimizer of \S\ref{sec:bi-objective} over 10 benchmarks spanning 6 domains.  Search-space sizes range from 18 configurations on each natural\_plan subtask (3 methods $\times$ 6 models, enumerated completely below the optimizer's 20-configuration exhaustive-fallback threshold) to over 10{,}000 on MedCalc (8 genes); the remaining benchmarks were searched by NSGA-II over five generations of population 12.  Each configuration is evaluated on 50 validation cases; frontier configurations (those Pareto-optimal in cost--correctness on the validation set) are re-run on a held-out test split where one is available.

\begin{table*}[ht]
\centering
\small
\caption{Per-benchmark NSGA-II frontier summary. \emph{Frontier size} is the
number of Pareto-optimal configurations on the validation split; \emph{valid}
is the highest validation correctness on the frontier; \emph{test} is the same
configuration's correctness on the held-out test split. \emph{Cheapest cost}
is the lowest LLM cost per query (USD) on the validation frontier.
\emph{Frontier methods} lists the distinct top-level methods that appear on
the validation frontier, in descending order of contribution; method labels
follow Table~\ref{tab:opt-search-space}.  Test split is unavailable for
FinQA (the public release ships only the dev split, used here as
\emph{valid}).}
\label{tab:opt-cross}
\begin{tabular}{lcccccl}
\toprule
                 & frontier & \multicolumn{2}{c}{best correct (\%)} & $n_{\mathrm{test}}$ & cheapest        & frontier methods                              \\
\cmidrule(lr){3-4}
benchmark        & size     & valid           & test                & (cases)             & cost (\$/case)  & (deduped)                                     \\
\midrule
sports             & 2 & 96.0 & 94.0 & 100 & $2.7\!\times\!10^{-5}$ & \texttt{struct} \\
tabmwp             & 5 & 92.0 & 94.2 & 500 & $4.3\!\times\!10^{-5}$ & \texttt{struct, pot, wf} \\
rulearena\_nba     & 4 & 90.5 & 69.6 & 46  & $2.2\!\times\!10^{-3}$ & \texttt{r\_learn, wf, unstr} \\
musr\_object       & 4 & 86.0 & 78.0 & 50  & $3.3\!\times\!10^{-4}$ & \texttt{wf\_orch, struct, zs\_cot} \\
finqa              & 6 & 82.0 & ---  & --- & $1.4\!\times\!10^{-4}$ & \texttt{struct, wf, wf\_orch} \\
musr\_team         & 4 & 80.0 & 62.0 & 50  & $4.6\!\times\!10^{-4}$ & \texttt{pot, wf\_orch, zs\_cot} \\
medcalc            & 1 & 78.0 & 41.7 & 300 & $7.6\!\times\!10^{-4}$ & \texttt{struct} \\
musr\_murder       & 2 & 76.0 & 80.0 & 50  & $1.5\!\times\!10^{-4}$ & \texttt{struct} \\
natplan\_meeting   & 3 & 44.0 & 54.0 & 50  & $2.1\!\times\!10^{-4}$ & \texttt{struct} \\
natplan\_trip      & 4 & 22.0 & 22.0 & 50  & $1.5\!\times\!10^{-4}$ & \texttt{wf, struct} \\
\bottomrule
\end{tabular}

\end{table*}

The optimizer discovers a non-trivial Pareto frontier on every benchmark, ranging from a 6-point frontier on FinQA to a single-point degenerate one on MedCalc.  Validation rankings mostly survive on test: across the eight benchmarks with public test sets, the best-validation configuration loses more than 10pp on test on three (MedCalc, RuleArena/NBA, MuSR/team), and on five it loses less than 4pp or improves.

\begin{figure}[ht]
\centering
\includegraphics[width=0.9\textwidth]{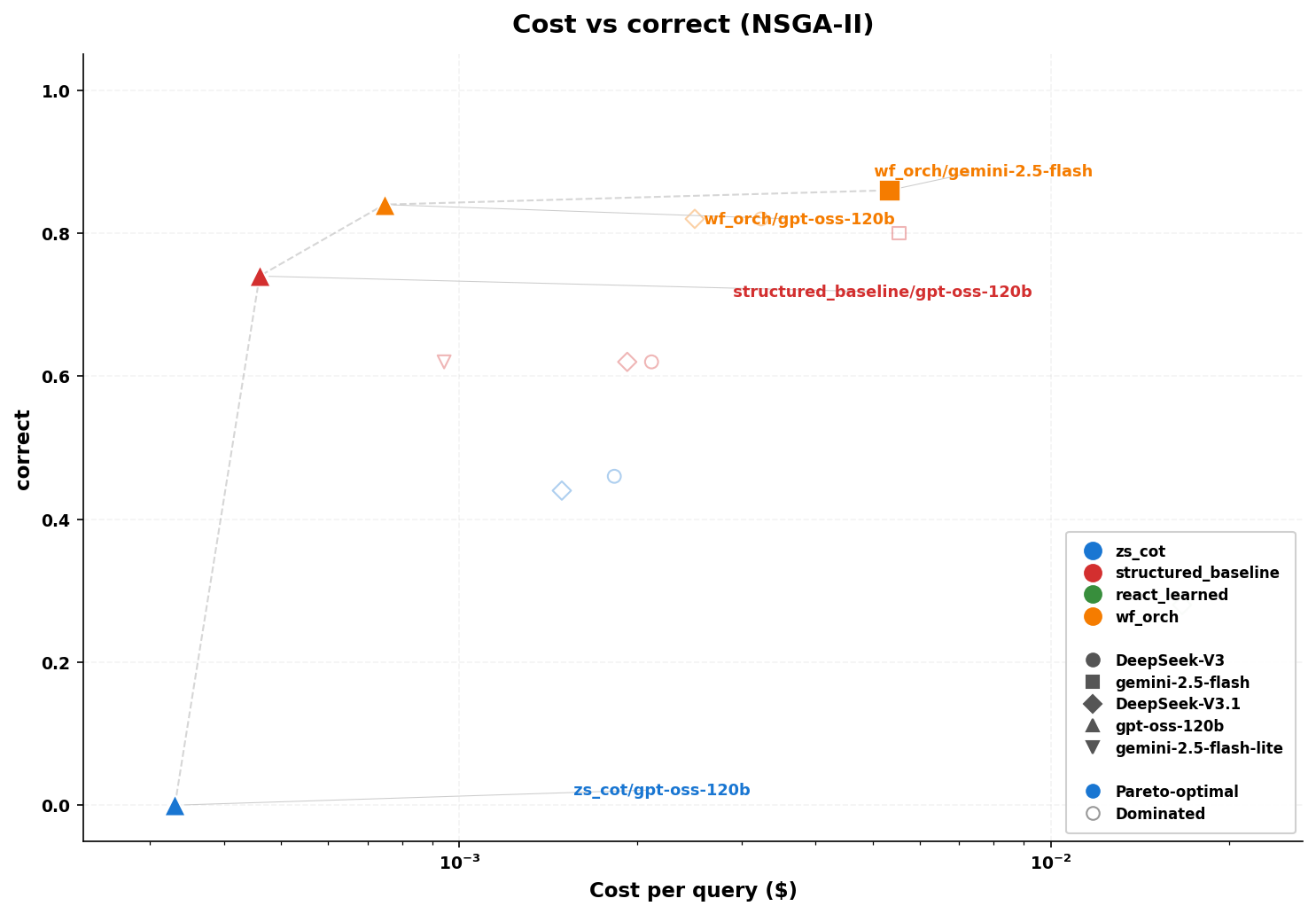}
\caption{NSGA-II Pareto frontier on MuSR/object\_placements (cost on the
validation split, USD per case, log axis, vs.\ correctness; $n_{\mathrm{valid}}=50$).
Filled markers are frontier-optimal; hollow markers are dominated samples;
colour encodes top-level method, marker shape encodes backend LLM.  The two
frontier points labelled \texttt{wf\_orch/gemini-2.5-flash} (86\% valid) and
\texttt{wf\_orch/gpt-oss-120b} (84\% valid) are the same seeded orchestration
workflow bound to two different backbones at roughly 1/7 cost ratio; both
score 78\% on the held-out test split.}
\label{fig:nsga2-musr-object}
\end{figure}

\textbf{Heterogeneous LLM assignment delivers cost-quality tiers through a single workflow.} The structural payoff of multi-objective optimization is visible in MuSR/object\_placements (Table~\ref{tab:opt-cross}, frontier of 4; Figure~\ref{fig:nsga2-musr-object}): two of the four frontier configurations are the seeded orchestration workflow of \S\ref{sec:learn-workflow}, one bound to gemini-2.5-flash at 86\% valid and the other to gpt-oss-120b at 84\% valid for roughly 1/7 the cost.  On the held-out test split both configurations score 78\%, so the validation spread vanishes while the cost spread remains; the frontier here is not a list of strictly better choices but a ladder of cost--quality tiers through the same learned workflow.  Per-interface model assignments on the discovered frontiers, and the corresponding per-benchmark sample distributions, are tabulated in Appendix~\ref{app:opt-details}.

\textbf{The optimizer's frontiers track the per-benchmark verdicts of the learning experiments.} Learned components are \textbf{retained} where the corresponding learner reported \textbf{headroom over hand-engineering}, and \textbf{pruned} where it did not.  The seeded orchestration workflow of \S\ref{sec:learn-workflow} lands on the frontier of MuSR/object\_placements (twice, as above) and MuSR/team, two of the cases on which \S\ref{sec:learn-workflow-results} reports the orchestration learner improving over its iteration-zero baseline; it is sampled but pruned on Sports and TabMWP, where the hand-engineered workflow already saturates the model.  Induced pseudo-tool toolkits behave similarly: \texttt{react\_learned} occupies two of the four frontier points on RuleArena/NBA, at 90.5\% (DeepSeek-V3) and 74\% (gpt-oss-20b), reflecting the $+6.5$pp induction gain reported in \S\ref{sec:learn-ptools-results}.  The optimizer is \textbf{therefore not just a Pareto search but a verifier}: it confirms or contradicts each learner's per-benchmark claim.

\textbf{Caveats.} Three points qualify the cross-benchmark reading.

\emph{Sampling concentration.} NSGA-II concentrates effort on early-population winners: across the eight stochastic-search benchmarks, the most-sampled method received between 26\% and 62\% of all configuration evaluations, while methods such as the seeded-orchestration workflow received between 0 and 10.

\emph{Provider-format wart.} Under our Together AI deployment, DeepSeek V3 and V3.1 emit native special-token tool-call sequences that pydantic-ai never dispatches, zero-rating every (\texttt{react}, V3) and (\texttt{react}, V3.1) configuration; NSGA-II Pareto-prunes these after the first generation.\footnote{Appendix~\ref{app:opt-details} lists the affected (method, model) cells.}

\emph{Test generalisation.} MedCalc's best-validation configuration drops from 78\% valid to 41.7\% test (the largest gap in the table),\footnote{Both MedCalc/formulas and MedCalc/rules splits were combined and evaluated together in these experiments, as MedCalc} and on MuSR/team the valid-winning \texttt{pot} configuration tests at 62\% while \texttt{wf\_orch} on DeepSeek-V3.1, also on the frontier, tests at 70\%. So the optimizer's valid-best is not always the test-best frontier point.

\subsection{NSGA-II search-space details}\label{app:opt-search-space}

For each benchmark we declare the search space in a YAML file at \texttt{benchmarks/<bench>/nsga2*.yaml}.  Each declaration consists of (i)~the entry interface; (ii)~a shared model pool, fixed across our experiments at six LLMs (DeepSeek-V3, DeepSeek-V3.1, gpt-oss-20b, gpt-oss-120b, gemini-2.5-flash, gemini-2.5-flash-lite); (iii)~a top-level \emph{method gene} listing the alternative bindings for the entry interface, with each value expanding to the full set of dotlist overrides that bind the entry and any helper interfaces it requires; and (iv)~a per-sub-interface method gene where applicable.  The encoder treats top-level method as the primary categorical gene; the model gene and sub-interface genes contribute additional positions to the chromosome (\texttt{src/secretagent/optimize/encoder.py}).  Search spaces range from 18 configurations on each natural\_plan subtask (which the optimizer enumerated below the 20-configuration exhaustive-fallback threshold) to over 10{,}000 on MedCalc, with RuleArena/NBA at 360 (\S\ref{sec:bi-objective}).

\begin{table*}[ht]
\centering
\small
\caption{Per-benchmark top-level method gene options.  Method labels: \texttt{struct} (\textsf{structured\_baseline}; default LLM call against the entry interface), \texttt{unstr} (\textsf{unstructured\_baseline}; zero-shot prompt), \texttt{wf} (\textsf{workflow}; hand-engineered Python orchestration over the toolkit), \texttt{pot} (\textsf{program-of-thought}; sandboxed code generation), \texttt{react} (\textsf{simulate\_pydantic} ReAct loop over the engineered toolkit), \texttt{react\_learned} (ReAct loop over the induced ptool toolkit of \S\ref{sec:learn-ptools-results}), \texttt{wf\_orch} (the seeded orchestration workflow of \S\ref{sec:learn-workflow}), and \texttt{zs\_cot} (zero-shot chain-of-thought prompt template).  Sub-interface genes per benchmark are documented in the YAML files; the largest is MedCalc (3 sub-interfaces: \textsf{identify\_calculator}, \textsf{extract\_clinical\_values}, \textsf{compute\_calculation}).}
\label{tab:opt-search-space}
\begin{tabular}{ll}
\toprule
benchmark              & top-level method options                                                            \\
\midrule
sports                 & \texttt{struct, unstr, wf, react, react\_learned, wf\_orch}                          \\
finqa                  & \texttt{struct, unstr, wf, pot, react, wf\_orch}                                     \\
tabmwp                 & \texttt{struct, wf, pot, react, wf\_orch}                                            \\
rulearena\_nba         & \texttt{struct, unstr, wf, pot, react, react\_learned}                               \\
musr/$\{$murder, object, team$\}$ & \texttt{struct, zs\_cot, wf, pot, react, react\_learned, wf\_orch}                   \\
medcalc                & \texttt{struct, wf, pot, react, react\_learned, wf\_orch}                            \\
natural\_plan/$\{$meeting, trip$\}$ & \texttt{struct, zs\_cot, wf}                                                          \\
\bottomrule
\end{tabular}

\end{table*}

\subsection{Cache effectiveness in NSGA-II}\label{app:opt-cache}

The body claim that caching makes optimization affordable is load-bearing for the practicality of the framework: an uncached NSGA-II sweep at our search-space sizes (Table~\ref{tab:opt-search-space}) and per-configuration sample size ($n_{\mathrm{valid}}=50$) would cost several times the cached version.  We measured this directly on RuleArena/NBA, the only run for which we recorded both the cache-hit cost and the equivalent no-cache cost (the latter computed by summing the per-call \texttt{cost} field that each evaluator records regardless of cache state).

Across the 43-configuration NBA sweep, the cache absorbed \textbf{87.5\%} of the run's notional API spend: the run cost \$11.90 in fresh API calls against a \$95.44 no-cache equivalent, with cache hits contributing \$83.54.  This is no surprise; NSGA-II revisits good configurations as the population converges, and similar chromosomes call the same sub-interfaces with the same inputs.  At the configuration level (Table~\ref{tab:cache-by-gen}) the hit rate climbs from 0\% in generation 0 to 58\% by generation 4 and plateaus; sub-call hits within configurations contribute the difference between this 0--58\% per-configuration range and the aggregate 87.5\% figure.\footnote{Cache keys are \texttt{(prompt, model)} only.  Hyperparameter sweeps over the same prompt (e.g.\ varying \texttt{max\_tokens} or \texttt{reasoning\_effort}) require explicit invalidation via \texttt{cachier.enable\_caching=false}.}

\begin{table*}[ht]
\centering
\small
\caption{Per-generation NSGA-II cache effectiveness at the configuration level on RuleArena/NBA.  ``New evaluations'' is the number of unique chromosomes the optimizer proposed and evaluated in each generation; ``cache hits'' is the number it proposed and recognised as already-evaluated; ``hit \%'' is the fraction of considered chromosomes that were already-seen.  The aggregate sub-call cache effectiveness over the entire run is 87.5\% of notional API spend, higher than any per-generation row above because it counts ptool-level call repetition within configurations as well as configuration-level repetition across generations.}
\label{tab:cache-by-gen}
\begin{tabular}{ccccc}
\toprule
generation & new evaluations & cache hits & considered & hit \% \\
\midrule
0 & 12 &  0 & 12 &  0.0 \\
1 & 10 &  1 & 11 &  9.1 \\
2 &  7 &  3 & 10 & 30.0 \\
3 &  5 &  6 & 11 & 54.5 \\
4 &  5 &  7 & 12 & 58.3 \\
5 &  4 &  5 &  9 & 55.6 \\
\bottomrule
\end{tabular}

\end{table*}




\subsection{Per-benchmark optimization details}\label{app:opt-details}

\begin{figure}[ht]
\centering
\includegraphics[width=0.7\textwidth]{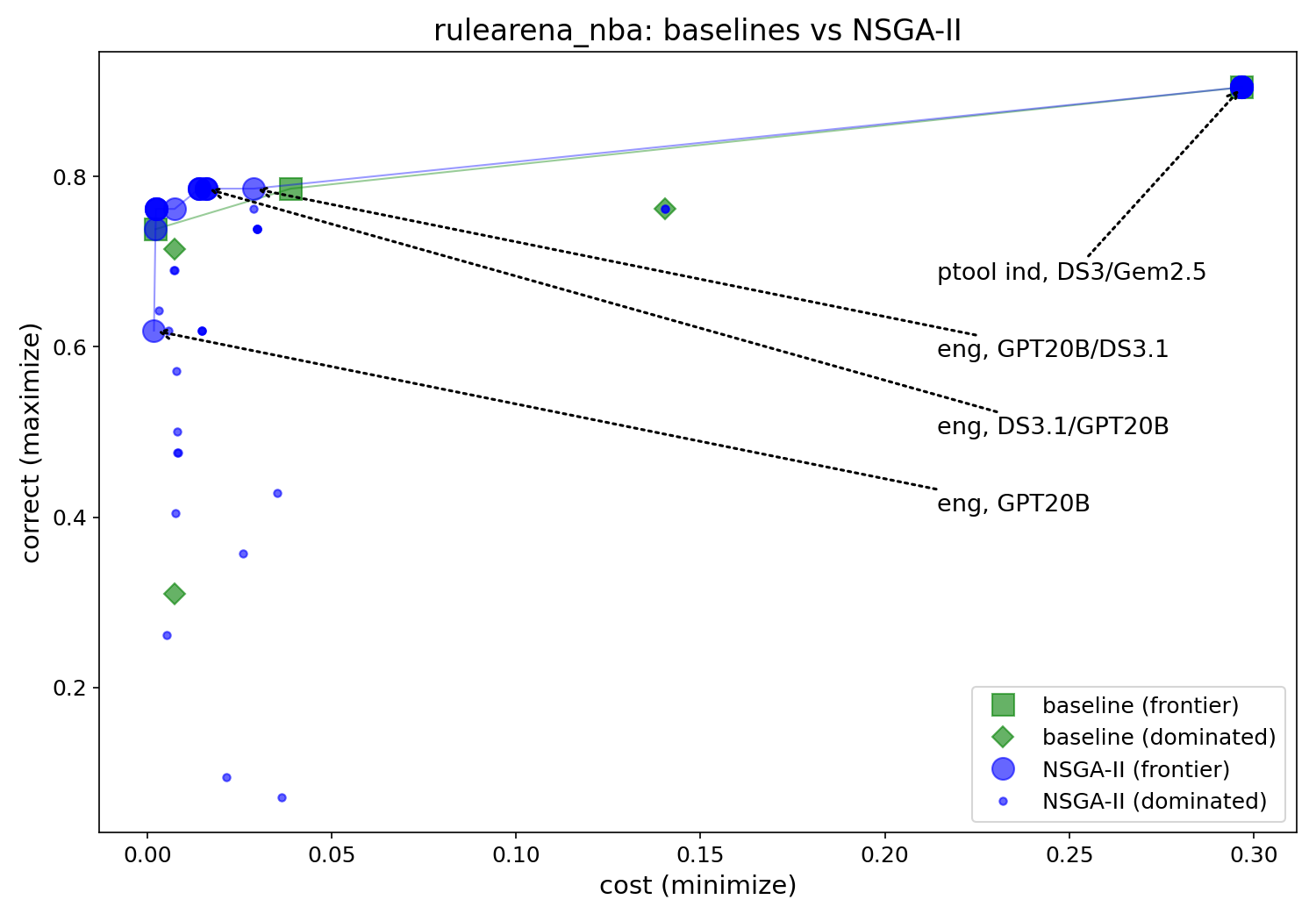}
\caption{RuleArena/NBA: engineered six-model \texttt{react\_learned} baseline
(green) versus the NSGA-II frontier (blue); validation cost (USD/case,
linear axis) versus correctness. Squares mark frontier-optimal, diamonds
mark dominated. Point labels show each configuration's top-level method and
model assignment(s). NSGA-II recovers the high-accuracy DeepSeek-V3
baseline (top-right) and finds two cheaper frontier configurations the
engineered sweep did not produce.}
\label{fig:nba-baseline-vs-nsga}
\end{figure}

This section provides per-benchmark detail supporting the cross-benchmark claims of \S\ref{sec:opt-results}: NSGA-II method-gene sample frequencies (Table~\ref{tab:opt-method-freq}), Pareto-frontier configurations evaluated on the held-out test split (Table~\ref{tab:opt-test-pass}), the affected (method, model) cells of the V3/V3.1 ReAct provider-format wart (Table~\ref{tab:opt-v31-wart}), and a worked-example contrast for RuleArena/NBA between the optimizer's frontier and the six-model engineered baseline (Figure~\ref{fig:nba-baseline-vs-nsga}).

\begin{table*}[ht]
\centering
\small
\caption{Per-benchmark NSGA-II method-gene sample counts.  Cells show the
number of evaluated configurations that used each method on the validation
split.  ``---'' marks methods not in the benchmark's YAML gene set
(Table~\ref{tab:opt-search-space}); ``0'' marks methods that are in the gene
set but received zero NSGA-II samples within the population-12, 5-generation
budget.  The most-sampled method received between 26\% (musr\_team
\texttt{pot}) and 62\% (tabmwp \texttt{struct}) of total evaluations on each
benchmark; \texttt{wf\_orch} was sampled between 0 and 10 times on benchmarks
where its yaml entry exists.  MedCalc is omitted: sample counts on its
single-frontier sweep are not load-bearing.}
\label{tab:opt-method-freq}
\begin{tabular}{lrrrrrrrrr}
\toprule
benchmark              & total & struct & unstr & wf  & pot & react & r\_lrn & wf\_orch & zs\_cot \\
\midrule
sports                 & 50 & 22 & 10 &  7 & --- &  6 &  0 &  5 & --- \\
tabmwp                 & 42 & 26 & --- &  4 &  7 &  4 & --- &  1 & --- \\
finqa                  & 43 & 18 &  2 & 10 &  2 &  1 & --- & 10 & --- \\
musr\_murder           & 48 & 21 & --- &  6 &  5 &  0 &  9 &  0 &  7 \\
musr\_object           & 54 & 18 & --- &  5 &  3 &  0 &  5 & 10 & 13 \\
musr\_team             & 50 &  3 & --- &  8 & 13 &  1 &  9 &  4 & 12 \\
natplan\_meeting       & 18 &  6 & --- &  6 & --- & --- & --- & --- &  6 \\
natplan\_trip          & 18 &  6 & --- &  6 & --- & --- & --- & --- &  6 \\
rulearena\_nba         & 43 &  3 & 14 & 10 &  3 &  0 & 13 & --- & --- \\
\bottomrule
\end{tabular}

\end{table*}

\begin{table*}[ht]
\centering
\small
\caption{Per-benchmark Pareto-frontier configurations re-evaluated on the
held-out test split.  Each row is a validation-frontier configuration; valid
\% is the optimizer's reported correctness on the 50-case validation split;
test \% is the same configuration's correctness on the held-out test split
of size $n_{\mathrm{test}}$.  Within each benchmark, rows are ordered by
descending validation correctness.  FinQA is omitted because its public
release ships only the dev split (used here as \emph{valid}).  Method labels
follow Table~\ref{tab:opt-search-space}.}
\label{tab:opt-test-pass}
\begin{tabular}{llrrr}
\toprule
benchmark              & method / model                              & valid \% & test \%  & $n_{\mathrm{test}}$ \\
\midrule
sports                 & struct / gemini-2.5-flash                   & 96.0 & 94.0 & 100 \\
sports                 & struct / gemini-2.5-flash-lite              & 90.0 & 82.0 & 100 \\
\midrule
tabmwp                 & struct / gemini-2.5-flash                   & 92.0 & 94.2 & 500 \\
tabmwp                 & pot / gpt-oss-120b                          & 90.0 & 87.4 & 500 \\
tabmwp                 & wf / gpt-oss-120b                           & 88.0 & 74.0 & 500 \\
tabmwp                 & struct / gpt-oss-120b                       & 86.0 & 72.8 & 500 \\
tabmwp                 & struct / gemini-2.5-flash-lite              & 70.0 & 64.8 & 500 \\
\midrule
rulearena\_nba         & react\_learned / DeepSeek-V3                & 90.5 & 69.6 &  46 \\
rulearena\_nba         & wf / gemini-2.5-flash-lite                  & 78.6 & 54.3 &  46 \\
rulearena\_nba         & unstr / gemini-2.5-flash-lite               & 76.2 & 58.7 &  46 \\
rulearena\_nba         & react\_learned / gpt-oss-20b                & 73.8 & 50.0 &  46 \\
\midrule
musr\_object           & wf\_orch / gemini-2.5-flash                 & 86.0 & 78.0 &  50 \\
musr\_object           & wf\_orch / gpt-oss-120b                     & 84.0 & 78.0 &  50 \\
musr\_object           & struct / gpt-oss-120b                       & 74.0 & 64.0 &  50 \\
musr\_object           & zs\_cot / gpt-oss-120b                      &  0.0 &  2.0 &  50 \\
\midrule
musr\_team             & pot / DeepSeek-V3.1                         & 80.0 & 62.0 &  50 \\
musr\_team             & wf\_orch / DeepSeek-V3.1                    & 76.0 & 70.0 &  50 \\
musr\_team             & zs\_cot / DeepSeek-V3.1                     & 52.0 & 34.0 &  50 \\
musr\_team             & zs\_cot / gpt-oss-120b                      &  0.0 &  0.0 &  50 \\
\midrule
musr\_murder           & struct / gemini-2.5-flash                   & 76.0 & 80.0 &  50 \\
musr\_murder           & struct / gemini-2.5-flash-lite              & 56.0 & 64.0 &  50 \\
\midrule
medcalc                & struct / gpt-oss-120b                       & 78.0 & 41.7 & 300 \\
\midrule
natplan\_meeting       & struct / gpt-oss-20b                        & 44.0 & 54.0 &  50 \\
natplan\_meeting       & struct / DeepSeek-V3.1                      & 30.0 & 36.0 &  50 \\
natplan\_meeting       & struct / gemini-2.5-flash-lite              & 26.0 & 24.0 &  50 \\
\midrule
natplan\_trip          & wf / DeepSeek-V3.1                          & 22.0 & 22.0 &  50 \\
natplan\_trip          & struct / DeepSeek-V3                        & 20.0 & 18.0 &  50 \\
natplan\_trip          & struct / DeepSeek-V3.1                      & 18.0 & 18.0 &  50 \\
natplan\_trip          & struct / gemini-2.5-flash-lite              & 16.0 & 16.0 &  50 \\
\bottomrule
\end{tabular}

\end{table*}

\newpage

\paragraph{V3/V3.1 ReAct provider-format wart.} Under our Together AI deployment, DeepSeek V3 and V3.1 emit native special-token tool-call sequences (e.g.\ \texttt{<|tool\_calls\_begin|>...<|tool\_calls\_end|>}) rather than the OpenAI-style JSON \texttt{tool\_calls} field.  The pydantic-ai harness used by our \texttt{react} and \texttt{simulate\_pydantic} factories parses these as plain text and never dispatches the call, so the agent never reaches \texttt{finish}.  Every affected configuration returns $0\%$ accuracy at non-zero cost; NSGA-II Pareto-prunes them after one generation.  The same wart removes the (\texttt{wf\_orch}, V3) and (\texttt{wf\_orch}, V3.1) cells on MedCalc, where the orchestration learner ran in non-seed mode and exposed the evolved toolkit through \texttt{simulate\_pydantic} rather than as a directly-callable Python function.

\begin{table*}[ht]
\centering
\small
\caption{(method, model) cells affected by the V3/V3.1 ReAct provider-format
wart, by benchmark.  Cells in benchmarks not listed are unaffected (the
benchmark's YAML gene set does not include the method, or the method is not
dispatched through pydantic-ai).  natural\_plan/$\{$meeting, trip$\}$ are
both unaffected; their gene sets contain no \texttt{react}-family methods.}
\label{tab:opt-v31-wart}
\begin{scriptsize}
\begin{tabular}{ll}
\toprule
affected (method, model) cell                                                 & affected benchmarks                                       \\
\midrule
(\texttt{react}, V3), (\texttt{react}, V3.1)                                  & sports, tabmwp, finqa, musr/$\{$murder, object, team$\}$, rulearena\_nba, medcalc \\
(\texttt{react\_learned}, V3), (\texttt{react\_learned}, V3.1)                & sports, musr/$\{$murder, object, team$\}$, rulearena\_nba, medcalc                \\
(\texttt{wf\_orch}, V3), (\texttt{wf\_orch}, V3.1)                            & medcalc                                                  \\
\bottomrule
\end{tabular}
\end{scriptsize}
\end{table*}

\newpage 

\paragraph{Worked example: NSGA-II versus engineered baselines on RuleArena/NBA.}
Figure~\ref{fig:nba-baseline-vs-nsga} contrasts the optimizer's Pareto frontier with a six-model engineered baseline (the \texttt{react\_learned} agent run separately on each model, with no sub-interface model gene).  The engineered frontier (green squares) tops out at \$0.30 and 90\% accuracy on \texttt{ptool-induced} with DeepSeek-V3.  NSGA-II (blue circles) recovers a comparable 90\% point at the same cost \emph{and} populates the cheap end of the frontier with two further configurations the engineered sweep never explored: an \texttt{eng}/GPT-OSS-20B configuration at \$0.002 and 76\% accuracy, and an \texttt{eng}/DeepSeek-V3.1 configuration at \$0.014 and 79\% accuracy.  The optimizer's value, on this benchmark, is at the cheap end of the frontier; the high-accuracy point was already reachable by tuning the model on the engineered toolkit.

\end{document}